\documentclass[journal,twoside,web]{ieeecolor}
\usepackage{tmi}
\usepackage{cite}
\usepackage{amsmath,amssymb,amsfonts}
\usepackage{algorithmic}
\usepackage{graphicx}
\usepackage{textcomp}

\usepackage{bbding}
\usepackage{makecell}
\usepackage{multirow}
\usepackage{booktabs}
\usepackage{color, xcolor, colortbl}
\usepackage{soul} 

\makeatletter
\let\NAT@parse\undefined
\makeatother
\definecolor{redtext}{RGB}{225,55,35}
\definecolor{bluetext}{RGB}{90,118,197}
\definecolor{yellowtext}{RGB}{240,193,68}
\definecolor{tmiblue}{RGB}{0,126,215}
\definecolor{bggreen}{RGB}{200,229,179}
\definecolor{bgblue}{RGB}{199,218,232}
\definecolor{bggray}{RGB}{217,217,217}
\usepackage[colorlinks=true,linkcolor=red,citecolor=tmiblue,urlcolor=tmiblue]{hyperref}

\def\BibTeX{{\rm B\kern-.05em{\sc i\kern-.025em b}\kern-.08em
    T\kern-.1667em\lower.7ex\hbox{E}\kern-.125emX}}
\begin{document}
\title{DDaTR: Dynamic Difference-aware Temporal Residual Network for Longitudinal Radiology Report Generation}
\author{Shanshan Song, Hui Tang, Honglong Yang, Xiaomeng Li \IEEEmembership{Member, IEEE}
\thanks{Corresponding author: Xiaomeng Li}
\thanks{This work was supported by a research grant from the Joint Research Scheme (JRS) under the National Natural Science Foundation of China (NSFC) and the Research Grants Council (RGC) of Hong Kong (Project No. N\_HKUST654/24), as well as a grant from the RGC of the Hong Kong Special Administrative Region, China (Project No. R6005-24). Shanshan Song, Hui Tang, Honglong Yang, and Xiaomeng Li are with the Department of Electronic and Computer Engineering, Hong Kong University of Science and Technology, Hong Kong, SAR, China (email: ssongan@connect.ust.hk; eehtang@ust.hk; hyangdh@connect.ust.hk; eexmli@ust.hk)}}

\maketitle

\begin{abstract}
Radiology Report Generation (RRG) automates the creation of radiology reports from medical imaging, enhancing the efficiency of the reporting process. 
Longitudinal Radiology Report Generation (LRRG) extends RRG by incorporating the ability to compare current and prior exams, facilitating the tracking of temporal changes in clinical findings.
Existing LRRG approaches only extract features from prior and current images using a visual pre-trained encoder, which are then concatenated to generate the final report. However, these methods struggle to effectively capture both spatial and temporal correlations during the feature extraction process. Consequently, the extracted features inadequately capture the information of difference across exams and thus underrepresent the expected progressions, leading to sub-optimal performance in LRRG. 
To address this,  we develop a novel dynamic difference-aware temporal residual network (DDaTR). 
In DDaTR, we introduce two modules at each stage of the visual encoder to capture multi-level spatial correlations. The Dynamic Feature Alignment Module (DFAM) is designed to align prior features across modalities for the integrity of prior clinical information. Prompted by the enriched prior features, the dynamic difference-aware module (DDAM) captures favorable difference information by identifying relationships across exams. 
Furthermore, our DDaTR employs the dynamic residual network to unidirectionally transmit longitudinal information, effectively modeling temporal correlations. Extensive experiments demonstrated superior performance over existing methods on three benchmarks, proving its efficacy in both RRG and LRRG tasks. Our code is published at \url{https://github.com/xmed-lab/DDaTR}

\end{abstract}


\section{Introduction}
\label{sec:introduction}

Radiology Report Generation (RRG) focuses on the automated creation of coherent and precise textual descriptions for radiology images, thereby reducing radiologists' workload \cite{esteva2019guide, sloan2024automated}. In a typical radiology workflow, radiologists not only deliver comprehensive analyses of clinical findings from the current exam, but also compare the current images with prior images and reports to identify significant progressions in key observations, assess treatment response, or detect early signs of disease recurrence.
\begin{figure}[t]
    \centering
    \includegraphics[scale=.32]{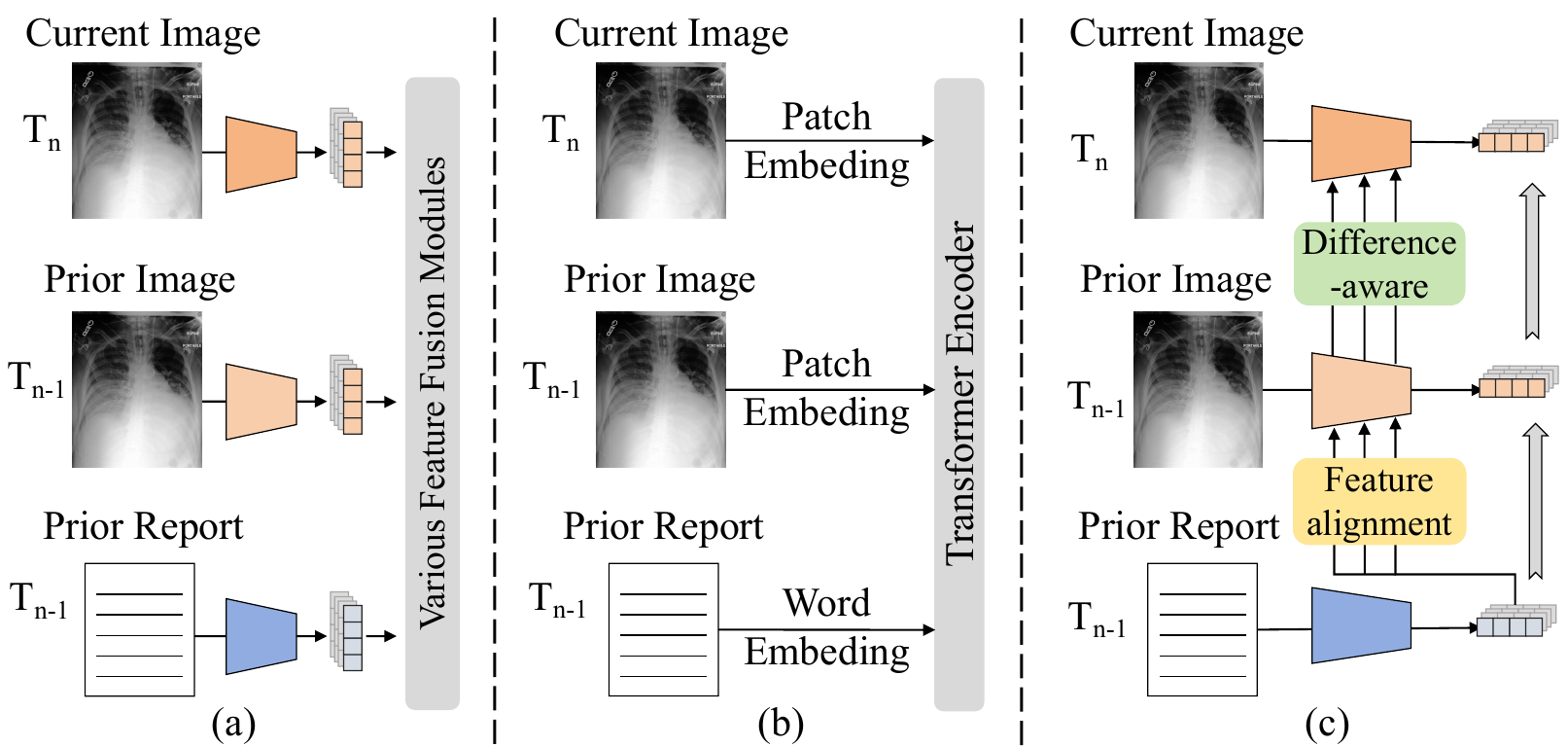}
    \caption{Comparison of our method with previous LRRG approaches. \textbf{(a)} represents methods that encode different modalities separately and subsequently fuse them using various feature fusion modules \cite{zhu2023Lr2gen, dalla2023controllable, wang2024hergen}. \textbf{(b)} denotes methods that embed longitudinal information into tokens, which are then processed collectively through a transformer-based encoder \cite{bannur2023Biovil, 24nicolson2023longitudinal, 25hou2023recap}. \textbf{(c)} is our proposed DDaTR, a novel approach for longitudinal information fusion that leverages prior image-text feature alignment and longitudinal difference-aware information.}
    \label{fig:introduction}
    \vspace{-6.5mm}
\end{figure}
To accurately generate the report with clinical progressions, Longitudinal Radiology Report Generation (LRRG) focuses on integrating a patient's current and prior exams to provide a comprehensive description of their condition in the report \cite{zhu2023Lr2gen}. Compared to RRG, this task monitors the progression of a patient's condition and can more effectively assist doctors in developing subsequent treatment plans \cite{zhu2023Lr2gen, dalla2023controllable, bannur2023Biovil}. Moreover, by leveraging historical clinical data, LRRG promotes diagnostic consistency across different visits and among different practitioners, reducing variability in interpretations. However, it is more challenging as it requires modeling spatio-temporal information to capture the core progression, and there is limited research addressing this problem. Consequently, our research is dedicated to addressing the challenges associated with the LRRG task.

\begin{figure}[htbp]
    \centering
    \includegraphics[scale=.33]{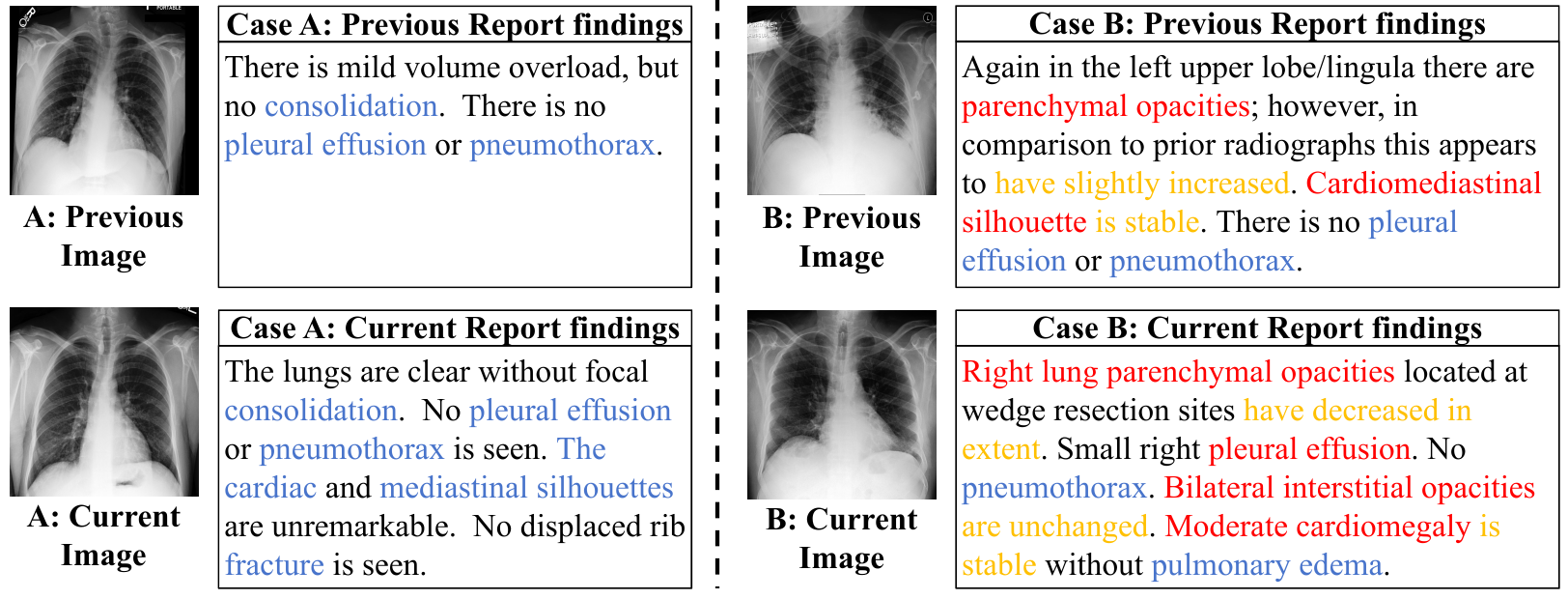}
    \caption{Different LRRG task examples. \textcolor{bluetext}{\textbf{Blue}} text indicates `negative observations’, \textcolor{redtext}{\textbf{red}} text indicates `positive observations’, and \textcolor{yellowtext}{\textbf{yellow}} text denotes the `progressions’. Both case A and case B include prior image and report for reference. However, in case A, there is no valuable longitudinal information to describe in the current report findings, indicating a minimal contribution from prior information. Conversely, case B primarily describes longitudinal information, demonstrating a significant contribution from prior information.}
    \label{fig:samplecase}
    \vspace{-6.5mm}
\end{figure}

Previous RRG works primarily adopt the encoder-decoder structure from the image captioning field \cite{06plummer2015flickr30k, 07vinyals2015show}. While image captioning typically targets short text descriptions, RRG requires the creation of fine-grained, long-text paragraphs for clinical findings. To achieve it, some studies improved the encoder-decoder framework by designing new networks and modules \cite{11chen2020R2gen, chen2021R2GenCMN, qin2022CMMRL, wang2023metransformer}. Some methods incorporated medical knowledge graphs or extra medical knowledge \cite{20liu2021auto, 22yang2022knowledge, 21huang2023kiut, li2023DCL}. Recently, several approaches have employed multi-task learning that facilitate the core generation task with other vision tasks, e.g., object detection to extract abnormal regions and anatomical structures \cite{17tanida2023RGRG}, classification to predict observation labels \cite{15hou2023organ, 16jin2024promptmrg}. Multi-task learning has proven effective in capturing fine-grained information and enhancing interpretability. However, above RRG approaches fail to generate longitudinal descriptions that align accurately with radiologists’ intentions. 

To address the challenge of LRRG, several approaches have been proposed to fuse the multi-period and multimodal features before generating the final reports. As shown in Fig. \ref{fig:introduction} (a), one category of works \cite{zhu2023Lr2gen, dalla2023controllable, wang2024hergen} applies various feature fusion modules, such as cross-attention modules,  after encoding each image or text with their respective pre-training encoders, and then feeds the fused features into a decoder to generate the corresponding reports. As shown in Fig. \ref{fig:introduction} (b), another category \cite{bannur2023Biovil, 24nicolson2023longitudinal, 25hou2023recap} leverages the powerful encoding capabilities of transformers to embed multi-period inputs, which uses modality-type and temporal encoding to differentiate tokens from different modalities and periods before decoding the report. Moreover, RECAP \cite{25hou2023recap} uses extra labels from the Chest ImaGenome dataset \cite{wu2chestImaGenome} to help learn the progressions across different periods during the encoding process.

However, the direct application of general visual pretraining encoders to the LRRG task for longitudinal feature extraction presents two key limitations, which we summarize as follows: 
(1) \textbf{Inaccurate perception of temporal correlations.} Existing methods overlook the different contributions of prior exams to current reports. As an example shown in Fig. \ref{fig:samplecase}, even when prior images and reports are available, it is not always necessary to include longitudinal information in the current report findings. When prior exam is unavailable, the model should rely solely on current-phase features. Moreover, the unidirectional nature of temporal information transmission only from prior to current is also neglected. When extracting visual features from the prior image, the model should consider the prior report only, without incorporating any information from the current examination.
(2) \textbf{Inaccurate perception of spatial correlations.} In this work, spatial correlations refer to the relationships among local-level features within a single X-ray image, as well as the correspondences of detailed textures and structural patterns across a longitudinal image pair. For single image, previous methods directly extract prior image features without aligning them with the corresponding reports, which may contain clinical prior knowledge. Due to this misalignment, the extracted visual features fail to accurately perceive known observations or abnormalities.
Additionally, traditional fusion methods independently encode each image and then fuse them afterward, disregarding the multi-level spatial relationships between longitudinal images during the encoding process which particularly neglects the relationship of low-level details and textures. 

To address the above limitations, we design a novel dynamic difference-aware temporal residual network (DDaTR). Addressing limitation (1), our entire network dynamically integrates temporal information from prior image features at different levels of abstraction into current image features. The residual connections are used to unidirectionally transmit temporal information, which is more suitable for modeling longitudinal image-text relationships by adaptively learning the contributions of different prior features. Furthermore, this unidirectional information flow aligns better with clinical practice, enhancing the focus on generating the current report. Addressing limitation (2), we introduce two modules at each stage of the visual encoder: the Dynamic Feature Alignment Module (DFAM) and the Dynamic Difference-aware Module (DDAM). DFAM employs the entire prior report's features to guide the prior image encoding, thereby better aligning prior image-text features. The DDAM leverages differences between two image features to adaptively enhance the current image feature, effectively addressing the insufficient perception of feature differences across multi-period images.

In summary, the contributions of our work are as follows:

\begin{itemize}
\item We propose a novel dynamic difference-aware temporal residual network (DDaTR) to better encode longitudinal information by adaptively learning the multi-level abstractions of different prior features and modeling the unidirectional information flow.

\item We introduce two plug-and-play modules: DFAM and DDAM. These modules can be integrated into any multi-scale visual encoder for longitudinal feature extraction. 

\item We conduct assessments not only on the standard MIMIC-CXR benchmark \cite{27johnson2019mimic} but also under various settings with different proportions of longitudinal reports, demonstrating the robustness and versatility of our approach across diverse clinical scenarios. Extensive experiments and analyses indicate that our method aligns more closely with clinical applications and achieves state-of-the-art results in both RRG and LRRG tasks.
\end{itemize}

\section{Related Work}

\subsection{Radiology Report Generation}

Recently, automated Radiology Report Generation (RRG) has attracted significant research interest. Most RRG studies use CNN-based or Transformer-based visual pre-training backbones to extract visual features and RNN-based \cite{10yang2021writing, xue2018rnn1, 09xie2019attention} or Transformer-based structures \cite{11chen2020R2gen, 12miura2021improving, 13liu2021exploring} to decode the corresponding reports. RRG primarily focuses on the characteristics of radiographic images and reports which present more challenges compared to image captioning tasks.

To generate long reports with detailed abnormal findings and enhanced cross-modal alignment, several studies have focused on improving the encoder-decoder framework by designing new networks and incorporating various modules. For instance, R2Gen \cite{11chen2020R2gen}, R2GenCMN \cite{chen2021R2GenCMN}, CMM-RL \cite{qin2022CMMRL} and METransformer \cite{wang2023metransformer}. Although these methods design new networks to generate more accurate and coherent radiology reports, the end-to-end image-to-text generation framework lacks fine-grained clinical guidance and explicit domain knowledge. This limitation hinders the clinical accuracy of the generated reports, resulting in a low Clinical Efficacy (CE) score. Considering the medical domain knowledge and severe visual and textual bias in RRG, several works leverage medical knowledge graphs or extra medical prior knowledge to aid in generating accurate radiology reports, such as KGAE \cite{20liu2021auto}, KMGen \cite{22yang2022knowledge}, KiUT \cite{21huang2023kiut} and DCL \cite{li2023DCL}. However, constructing high-quality knowledge graphs is both costly and labor-intensive. Furthermore, the above methods that rely on specific prior knowledge often fail to generalize effectively across new modalities or datasets.

In RRG, accurately describing the presence or absence of clinical observations is more crucial than the linguistic style or expression used. To capture more fine-grained clinical features and enhance clinical efficacy, many approaches employ multi-task learning methods. These methods not only support the core report generation task but also integrate other vision-related tasks to aid the accurate report generation. RGRG \cite{17tanida2023RGRG} utilizes the object detection task to extract abnormal regions and anatomical structures from the image, and then feeds them to a final report decoder. ORGAN \cite{15hou2023organ} leverages the classification task to predict observation labels and attributes of diseases from images, and then uses these information to enhance the decoding process. However, these methods employ fine-grained information in an implicit way, facing challenges in effectively establishing a correspondence between intermediate features and the report. To overcome this, PromptMRG \cite{16jin2024promptmrg} uses an extra disease classification branch, and then generate final report with the guidance of diagnosis-aware prompts. Due to make use of diagnostic prompts to explicitly guide the generation process, PromptMRG has better explainability and meets the actual clinical workflow. Therefore, we choose the framework with diagnosis-aware prompts as our strong baseline for report generation. 

\subsection{Longitudinal Radiology Report Generation}

Previous RRG methods that focus on single-period data overlook historical records, which can result in inaccuracies and hallucinations about unobserved priors when training language models for report generation \cite{ramesh2022hallucination}. Recently, to improve the generation of reports that accurately reflect clinical progressions, various approaches have been developed to integrate multi-period and multimodal features, utilizing longitudinal information to produce more accurate final reports.

\begin{figure}[ht]
\centering
\includegraphics[scale=.6]{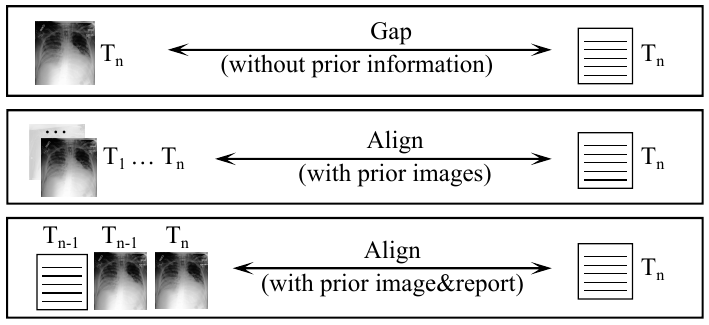}
\caption{Characteristics of Longitudinal Report Generation. The top panel highlights the gap between the current image and the corresponding report due to the lack of prior information. The middle panel illustrates how the final radiology report is constructed by integrating images from $T_1$ to $T_n$, taking into account both the patient’s current condition and their historical progressions in the real-world. The bottom panel shows the longitudinal information from $T_{n-1}$ report, $T_{n-1}$ and $T_n$ images are also aligned with the target $T_n$ report which is used in our work.}
\label{fig:formulate}
\vspace{-6.5mm}
\end{figure}

One category of works utilizes various feature fusion modules, such as cross-attention modules and projection modules, to integrate multi-period visual and textual features after encoding each image or text with their respective pre-training encoders, and then feeds the fused features into a decoder to generate the corresponding reports. In this study \cite{dalla2023controllable}, Faster R-CNN is employed to extract anatomical regions from both current and prior images. Subsequently, a longitudinal projection module is utilized to align, concatenate, and project these region representations into a cohesive joint representation. LR2Gen \cite{zhu2023Lr2gen} captures information from longitudinal patient visit records using a cross-attention-based multimodal fusion module and a hierarchical memory-driven decoder. HERGen \cite{wang2024hergen} employs a group causal transformer to efficiently integrate longitudinal data across patient visits.

Another category leverages the powerful encoding capabilities of transformers to embed multi-period images and texts. BioViL-T \cite{bannur2023Biovil} utilizes a CNN–Transformer hybrid multi-image encoder to extract longitudinal visual features. CXRMate  \cite{24nicolson2023longitudinal} incorporates longitudinal information by using prior reports as prompts, which are fed into a transformer decoder. Additionally, RECAP \cite{25hou2023recap} utilizes extra progression labels to assist the visual transformer-based encoder in learning the progressions across different periods.

\begin{figure*}[t]
    \centering
    \includegraphics[scale=.46]{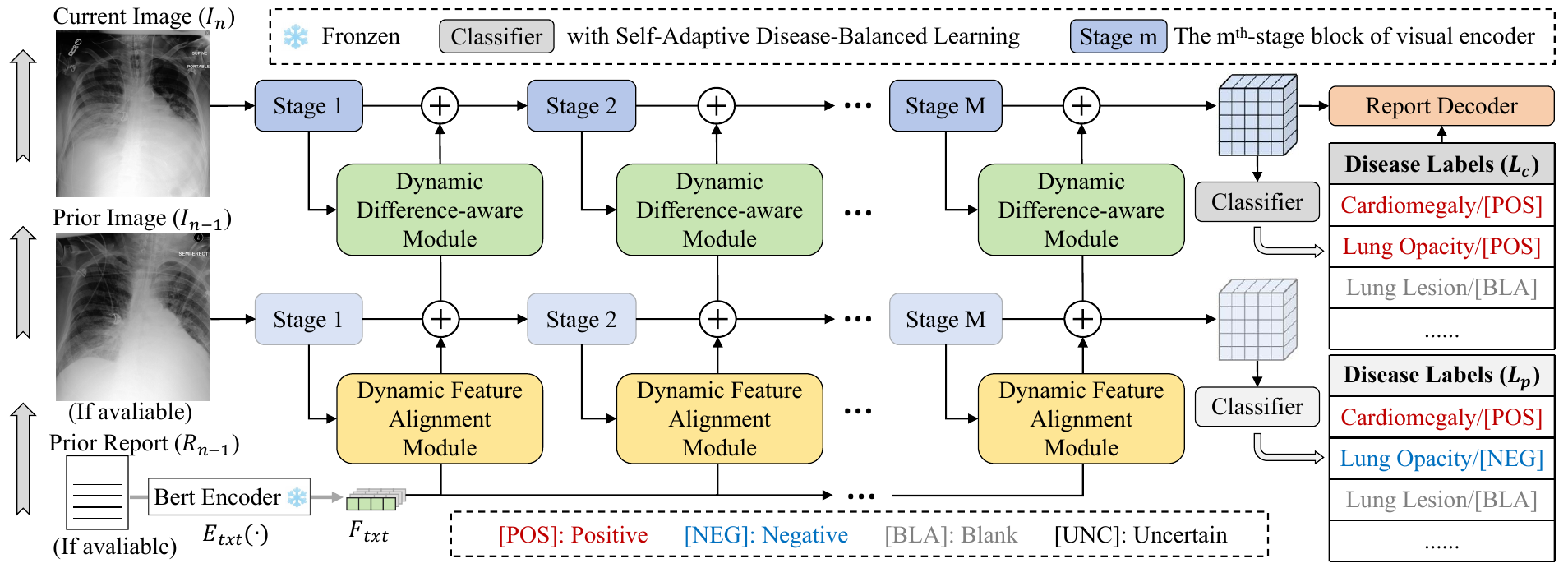}
    \caption{The framework of our proposed DDaTR. First, the prior report is processed through the text encoder $E_{txt}$ to extract textual features $F_{txt}$. Next, the current image, prior image, and $F_{txt}$ from the prior report are input into the DDaTR which incorporates the Dynamic Difference-aware Module and Dynamic Feature Alignment Module at multi-level encoding stage. The output features for the prior image and the current image are separately fed into different classifiers to obtain the respective disease labels. Finally, the current disease labels are used as prompts and input into the decoder, where the current image features are interactively fused with the decoder via cross-attention, ultimately generating the current report.}
    \label{fig:overall_network}
    \vspace{-5.5mm}
\end{figure*}

Although adding longitudinal features could alleviate the hallucinations about unobserved priors when training language models for report generation, there are still limitations in capturing the important longitudinal correlations. Firstly, they overlook the different contributions of previous image content (e.g., different viewpoints, limited longitudinal information or varying time intervals between examinations). This limitation leads to insufficient global temporal correlations. Secondly, previous feature fusion is used after visual encoder. Specifically, they use general visual pre-training encoders (e.g.,ResNet \cite{he2016deep}, SwinTransformer \cite{liu2021swin}) to extract visual features of each images independently, and then they leverage cross-attention based modules or transformer blocks to encode longitudinal features. However, the encoding process does not consider the relationships between multi-level features of longitudinal images, especially the low-level features including important details and texture information. This limitation leads to inaccurate local temporal correlations.
Thirdly, previous methods fail to account for the prior knowledge in previous report during the visual feature extraction. This report documents the patient’s clinical findings and diagnoses, reflecting the underlying clinical intent of the examination, which is important for perceiving prior visual semantics.

\section{Proposed Method}

To address the challenges of the LRRG task, previous approaches primarily adopted two types of settings. The first setting focused exclusively on generating longitudinal reports \cite{zhu2023Lr2gen, dalla2023controllable}, utilizing datasets that only included cases with prior records and disregarding data with only a single exam record. The second setting considered both single-period and multi-period data \cite{bannur2023Biovil, 25hou2023recap}. Their methodologies aimed to enhance the overall performance of general radiology report generation tasks, rather than focusing solely on longitudinal reports.

We believe that the incorporation of prior information is intended to improve the quality of the final report generation. If it proves effective only on longitudinal data but fails to surpass the performance of single-period report generation methods, the significance of incorporating prior data is limited. Therefore, we opt for the second setting. Our goal is to refine the utilization of multi-period information so that the performance of report generation is not only robust on multi-period datasets but also achieves state-of-the-art results in both single-period and multi-period report generation tasks.

\subsection{Problem Definition and Our Baseline Framework}

\textbf{Problem Definition}. We follow the setting in Recap \cite{25hou2023recap}, which involves generating a current report based on both the current image and, when available, prior image and report. However, our approach differs from Recap, which relies on the Chest ImaGenome dataset \cite{wu2chestImaGenome} that selects only Anteroposterior (AP) or Posteroanterior (PA) scans as priors and excludes lateral views. We obtain the longitudinal scans of each patient by chronologically organizing different studies according to corresponding timestamps. Our setting includes consideration of lateral views as well. As shown in Fig.\ref{fig:formulate}, the work in this paper is based on the hypothesis that longitudinal information from the $T_{n-1}$ report, $T_{n-1}$ image and $T_n$ image is aligned with the target $T_n$ report.

Specifically, define the current exam as the $n^{th}$ exam. The input consists of the current image $I_n \in \mathbb{I}^{ 3 \times 224 \times 224}$ , the prior image $I_{n-1} \in \mathbb{I}^{ 3 \times 224 \times 224}$, and the prior report $R_{n-1} = \{ r_1, r_2, ... r_{l_{n-1}} \}$. The output is the current report $R_n = \{ r_1, r_2, ... r_{l_n} \}$. $l_{n-1}$ and $l_n$ are the word length of prior and current report respectively. If $n = 1$, indicating that this is the patient's first exam, the $I_{n-1}$ and $R_{n-1}$ is None. 

\textbf{Our Baseline Framework}. The generation framework presented in our paper is based on PromptMRG \cite{16jin2024promptmrg}, which integrates an additional classification branch for generating diagnostic prompts used in producing the final report. In the disease classification branch, classification labels are derived using CheXbert \cite{smit2020CheXbert}, which converts reports into 14 predefined diseases. Each disease is described by one of four attributes: blank, positive, negative, or uncertain. Specifically, the encoder $E_{img}(\cdot)$ is firstly used to extract the visual feature $F_{img}$ from the input image $I_n$. 
\begin{equation}
\small
    F_{img} = E_{img} \left( I_n \right)
  \label{eq_PF1}
  \vspace{-2mm}
\end{equation}
Then, the visual feature $F_{img}$ is passed through a classifier $f_{cls}(\cdot)$ with self-adaptive disease-balanced learning (SDL) \cite{16jin2024promptmrg}, which outputs 14 predefined disease labels $L_{label} \in \mathbb{R}^{14}$. 
\begin{equation}
\small
    L_{label} = f_{cls} \left( F_{img} \right)
  \label{eq_PF2}
\end{equation}
Subsequently, these disease labels are converted into corresponding prompt tokens $T_{label}$, with each token corresponding to a specific disease. For each disease, one of the four new tokens, including [BLA], [POS], [NEG], and [UNC], is utilized to represent the disease attribute. Finally, both the prompt tokens $T_{label}$ and the visual feature $F_{img}$ are fed into a decoder $D(\cdot)$ to generate the target report $R_n$.
\begin{equation}
\small
    R_n = D \left( T_{label} , F_{img} \right)
  \label{eq_PF3}
\end{equation}
To perform LRRG based on the above baseline, we propose Dynamic Difference-aware Temporal Residual Network (DDaTR) $E_{ddatr}(\cdot)$ to integrate longitudinal multimodal information to get better visual features. Therefore, the encoding process in our paper is formalized as follows: 
{\small
\begin{equation}
    F_{img} = \begin{cases} 
    E_{ddatr} \left( I_n \right) & \text{if } n = 1 \\
    E_{ddatr} \left( I_n, I_{n-1}, R_{n-1} \right),  & \text{if } n > 1
    \end{cases}
    \vspace{-1mm}
  \label{eq_PF4}
\end{equation}}
\noindent where $F_{img}$ represents the output longitudinal visual feature, which differs from the baseline visual feature defined in Eqn. (\ref{eq_PF1}). After obtaining $F_{img}$, the subsequent step is the same as described in Eqn. (\ref{eq_PF2}) and Eqn. (\ref{eq_PF3}).

\subsection{Overview of Our Proposed DDaTR}
\label{sec:method_overall}

To better extract features from multimodal longitudinal inputs and effectively model the temporal relationships, we propose the Dynamic Difference-aware Temporal Residual Network (DDaTR). As shown in Fig. \ref{fig:overall_network}, our framework employs the encoder-decoder architecture with a disease classification branch. In this framework, the encoder $E_{ddatr}(\cdot)$ extracts the longitudinal features $F_{img}$ from the current image $I_n$, the prior image $I_{n-1}$, and the prior report $R_{n-1}$. Specifically, we begin by encoding the prior report $R_{n-1}$ using the pretrained BERT \cite{devlin2019bert} model $E_{txt}(\cdot)$ to extract the text feature $F_{txt}$. 
\begin{equation}
\small
    F_{txt} = E_{txt} \left( R_{n-1} \right)
  \label{eq_PF5}
\end{equation}
During the training process, the parameters of $E_{txt}(\cdot)$ are frozen without gradient updates. This not only reduces the number of training parameters but also focuses model optimization on the visual features. The text feature $F_{txt}$ is employed to align with the prior image features at each stage of the visual encoder, thereby enhancing the learned visual features of $I_{n-1}$ and facilitating the following perception of differences between two-period images. 

\begin{figure}[t]
\centering
\includegraphics[scale=.38]{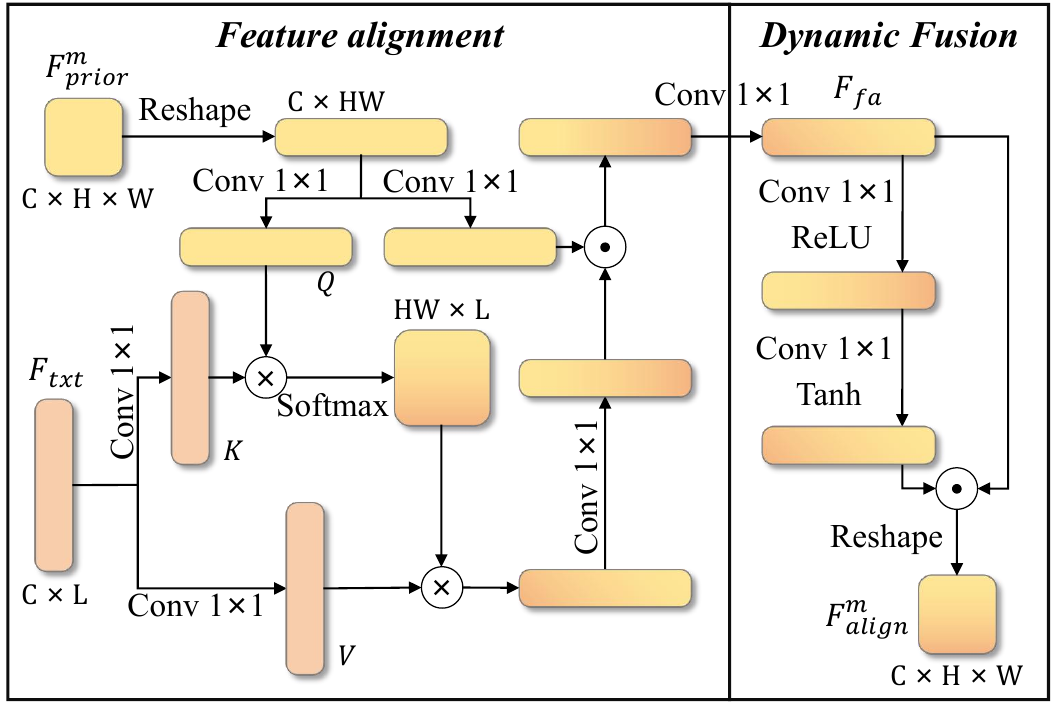}
\caption{The structure of Dynamic Feature Alignment Module (DFAM). Given prior image features $F_{prior}^{m}$ and prior text features $F_{txt}$ as inputs, the module processes these through feature alignment and dynamic fusion sequentially. Finally, the output is visual features $F_{align}^{m}$ that are aligned with the text and dynamically fused.}
\label{fig:DFAM}
\vspace{-5.5mm}
\end{figure}

Firstly, the output prior visual feature from the $m^{th}$ stage of the visual encoder is denoted as $F_{prior}^{m}$. $m \in [1,...,M]$, where $M$ is the number of stages of the visual encoder:
\begin{equation} 
\small
    F_{prior}^{m} = \begin{cases} 
    E_{prior}^{m} \left( I_{n-1} \right) & \text{if } m = 1 \\
    E_{prior}^{m} \left( \hat{F}_{prior}^{m-1} \right)  & \text{if } m > 1
    \end{cases}
  \label{eq_PF6}
\end{equation}
where $\hat{F}_{prior}^{m-1}$ represents the final output feature from the $m-1$ stage and $E_{prior}^{m}$ denotes the $m^{th}$ stage of prior visual encoder. For each stage of the prior visual encoder, we introduce the Dynamic Feature Alignment Module (DFAM) to dynamically fuse the image feature $F_{prior}^{m}$ and text feature $F_{txt}$, resulting in the aligned output feature $F_{align}^{m}$. 
\begin{equation}
\small
    F_{align}^{m} = DFAM \left( F_{prior}^{m}, F_{txt} \right)
  \label{eq_PF7}
\end{equation}
where the mechanics of the $DFAM(\cdot)$ are detailed in Sec. \ref{sec:DFAM}.
Next, this aligned output is added to $F_{prior}^{m}$ via a residual connection to obtain the final prior visual feature. The formula is given by:
\begin{equation}
\small
    \hat{F}_{prior}^{m} = F_{prior}^{m} + F_{align}^{m}
  \label{eq_PF8}
\end{equation}
where $\hat{F}_{prior}^{m}$ represents the final prior visual feature of the prior $m^{th}$ stage, which is subsequently fed into the next stage.

\begin{figure}[t]
\centering
\includegraphics[scale=.38]{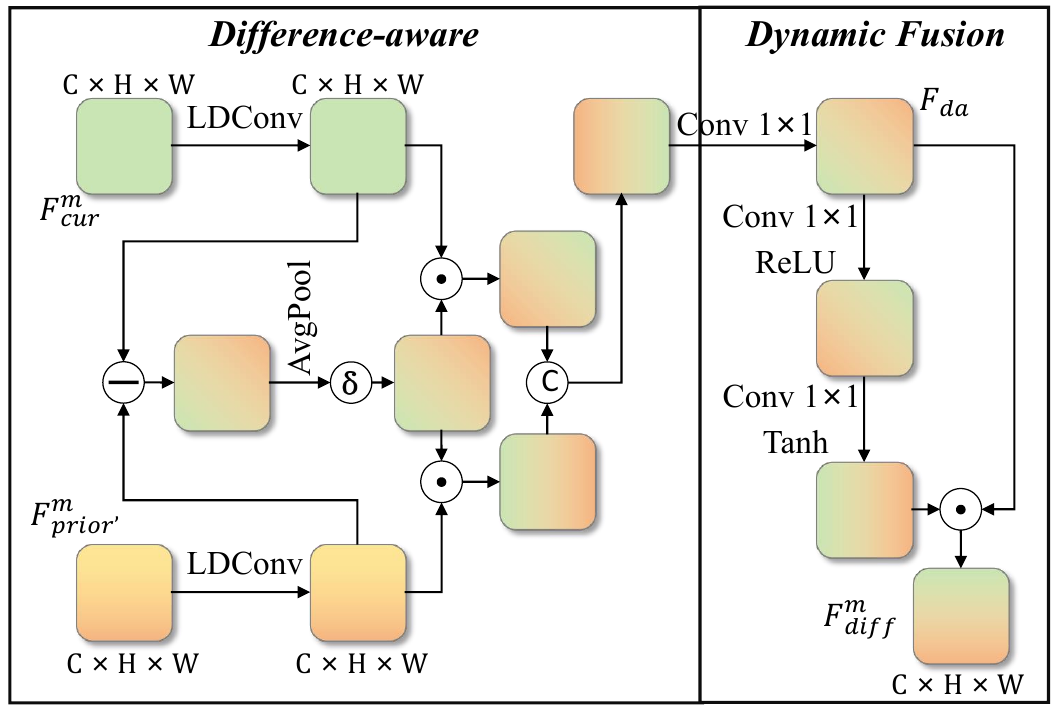}
\caption{The structure of Dynamic Difference-aware Module (DDAM). Given prior aligned visual features $\hat{F}_{prior}^{m}$ and current visual features $F_{cur}^{m}$ as inputs, the module processes these through difference-aware and dynamic fusion sequentially. Finally, the output is visual features $F_{diff}^{m}$ that are compared with the prior feature and dynamically fused.}
\label{fig:DDAM}
\vspace{-5.5mm}
\end{figure}

Secondly, the output current visual feature from the $m^{th}$ stage of current visual encoder is denoted as $F_{cur}^{m}$ as follows:
\begin{equation} 
\small
    F_{cur}^{m} = \begin{cases} 
    E_{cur}^{m} \left( I_n \right) & \text{if } m = 1 \\
    E_{cur}^{m} \left( \hat{F}_{cur}^{m-1} \right)  & \text{if } m > 1
    \end{cases}
  \label{eq_PF9}
\end{equation}
where $E_{cur}^{m}$ denotes the $m^{th}$ stage of current visual encoder and $\hat{F}_{cur}^{m-1}$ represents the final output feature from the current stage. Utilizing our proposed Dynamic Difference-aware Module (DDAM), we incorporate $F_{cur}^{m}$ with $\hat{F}_{prior}^{m}$, resulting in the difference-aware output $F_{diff}^{m}$. 
\begin{equation}
\small
    F_{diff}^{m} = DDAM \left( F_{cur}^{m}, \hat{F}_{prior}^{m} \right)
  \label{eq_PF10}
\end{equation}
The mechanics of the $DDAM(\cdot)$ are detailed in Sec. \ref{sec:DDAM}. This output is also added to $F_{cur}^{m}$ using a residual connection, subsequently feeding into the next stage. Ultimately, this process culminates in the acquisition of the current visual feature $\hat{F}_{cur}^{m}$. The equation can be expressed as:
\begin{equation} 
\small
    \hat{F}_{cur}^{m} = F_{cur}^{m} + F_{diff}^{m}
  \label{eq_PF11}
\end{equation}
During training, the final output prior and current visual features of the last stage deliver to two different classifiers to get their corresponding observation label: $L_p$, $L_c$. Then the decoder $D$ is utilized to generate the current report $R_n$, conditioned on both the longitudinal visual features $\hat{F}_{cur}^{M}$ and the diagnosis-driven prompts derived from $L_p$.
\begin{equation}
\small
    R_n = D \left( L_c, \hat{F}_{cur}^{M} \right) 
  \label{eq_PF12}
\end{equation}
It is worth noting that the classifier for the prior is used as a regularizer to assist in learning the prior-aligned visual feature, which does not need to be sent to the decoder. This is because the classifier for the prior part intended to convey additional supervision signals from prior image-text pairs which are groundtruth, effectively acting as a regularizer to ease the training process. Specifically, (1) it guides the learning of prior-aligned visual features by supervising the extraction of prior visual features using prior observation labels. This encourages the model to learn disease-discriminative patterns from prior visits, enabling it to extract relevant and useful prior information when generating reports from current images.  (2) due to the current features are supervised using the current labels, this design ensures semantic consistency between the prior and current visual encoders, which supports effective feature alignment and fusion across stages. As a result, our framework is better able to model longitudinal dependencies, leading to more accurate and context-aware report generation.

\begin{table*}[ht]
\centering
\caption{Comparison with existing report generation methods on MIMIC-CXR, Longitudinal-MIMIC and IU-XRay. * indicates results evaluated by us. For the MIMIC-CXR and IU-Xray, we adopt the evaluation outlined by RECAP\cite{25hou2023recap}, using the Macro score for CE metrics. For the Longitudinal-MIMIC, we follow L-R2Gen \cite{zhu2023Lr2gen}, calculating the Micro score for CE metrics.}
\label{tab:main result}
\resizebox{0.8\textwidth}{!}{\begin{tabular}{cccccccccc}
\hline
\multirow{2}{*}{\textbf{Datasets}} & \multirow{2}{*}{\textbf{Methods}} & \multirow{2}{*}{\textbf{Year}} & \multicolumn{3}{c}{\textbf{Clinical Efficacy}} & \multicolumn{4}{c}{\textbf{Natural Language Generation}} \\
\cline{4-10}
 &  &  &\multicolumn{1}{c}{\textbf{Precision}} & \multicolumn{1}{c}{\textbf{Recall}} & \multicolumn{1}{c}{\textbf{F1}} & \multicolumn{1}{c}{\textbf{BLEU-1}} & \multicolumn{1}{c}{\textbf{BLEU-4}} & \multicolumn{1}{c}{\textbf{METEOR}} & \multicolumn{1}{c}{\textbf{ROUGE-L}} \\
\hline
\multirow{13}{*}{MIMIC-CXR} 
 & R2Gen \cite{11chen2020R2gen} & 2020 & 0.333	& 0.273	& 0.276 & 0.353 & 0.103 & 0.142 & 0.270 \\
 & R2GenCMN \cite{chen2021R2GenCMN} & 2021 & 0.344	& 0.275	& 0.278 & 0.353 & 0.106 & 0.142 & 0.278 \\
 & KGAE \cite{20liu2021auto} & 2021 & 0.389 & 0.362 & 0.355 & 0.369 & 0.118 & 0.153 & \underline{0.295} \\
 & CMM-RL \cite{qin2022CMMRL} & 2022 & 0.342 & 0.294 & 0.292 & 0.381 & 0.109 & 0.151 & 0.287 \\
 & KMGen \cite{22yang2022knowledge} & 2022 & \underline{0.458} & 0.348 & 0.371 & 0.363 & 0.115 & - & 0.284 \\
 & KiUT \cite{21huang2023kiut} & 2023 & 0.371 & 0.318 & 0.321 & 0.393 & 0.113 & 0.160 & 0.285 \\
 & DCL\cite{li2023DCL} & 2023 & \textbf{0.471} & 0.352 & 0.373 & - & 0.109 & 0.150 & 0.284 \\
 & METrans.\cite{wang2023metransformer} & 2023 & 0.364 & 0.309 & 0.311 & 0.386 & 0.124 & 0.152 & 0.291 \\
 & ORGAN \cite{15hou2023organ} & 2023 & 0.416 & 0.418 & 0.385 & 0.407 & 0.123 & \underline{0.162} & 0.293 \\
 & R2GenGPT\cite{r2gengpt} & 2023 & 0.392 & 0.387 & 0.389 & \underline{0.411} & \textbf{0.134} & 0.160 & \textbf{0.297} \\
 & RECAP \cite{25hou2023recap} & 2023 & 0.389 & \underline{0.443} & \underline{0.393} & \textbf{0.429} & \underline{0.125} & \textbf{0.168} & 0.288 \\
 & HERGen \cite{wang2024hergen} & 2024 & 0.415 & 0.301 & 0.317 & 0.395 & 0.122 & 0.156 & 0.285 \\
 & PromptMRG*\cite{16jin2024promptmrg} & 2024 & 0.454 & 0.370 & 0.389 & 0.398 & 0.112 & 0.157	& 0.268	\\
  \rowcolor{cyan!60!blue!20}
 & \textbf{DDaTR(Ours)} & 2024 & 0.438 & \textbf{0.465} & \textbf{0.441 (+0.048)} & 0.401 & 0.113 & \underline{0.162} & 0.272 \\
 \hline
 \multirow{7}{*}{Longitudinal-MIMIC} 
 & AoANet\cite{huang2019AoANet} & 2021 & 0.437	& 0.249	& 0.371 & 0.272 & 0.080 & 0.115 & 0.249 \\
 & R2Gen\cite{11chen2020R2gen} & 2020 & 0.500	& 0.305	& 0.379 & 0.302 & 0.087 & 0.124 & 0.259 \\
 & R2GenCMN\cite{chen2021R2GenCMN} & 2021 & 0.521	& 0.396	& 0.449 & 0.305 & 0.085 & 0.126 & 0.265 \\
 & CNN+Trans\cite{moon2022cnntrans} & 2021 & 0.445 & 0.258	& 0.326 & 0.299 & 0.088 & 0.120 & 0.263 \\
 & LR2Gen\cite{zhu2023Lr2gen} & 2023 & \underline{0.538}	& 0.434	& 0.480 & 0.343 & 0.099 & 0.137 & \textbf{0.271}\\
 & PromptMRG*\cite{16jin2024promptmrg} & 2024 & 0.502 & \underline{0.543} & \underline{0.492} & \underline{0.390} & \underline{0.102} & \underline{0.152} & 0.263	\\
 \rowcolor{cyan!60!blue!20}
 & \textbf{DDaTR(Ours)} & 2024 & \textbf{0.539} & \textbf{0.584} & \textbf{0.527 (+0.035)} & \textbf{0.396} & \textbf{0.105} & \textbf{0.156} & \underline{0.266} \\
\hline
\multirow{7}{*}{IU-Xray} & R2Gen\cite{11chen2020R2gen}  & 2020 & 0.161 & 0.104 & 0.071 & 0.325 & 0.059 & 0.131 & 0.253 \\
& CVT2Dis.\cite{CVT2Dis} & 2023 & 0.325 & 0.166 & 0.155 & 0.383 & 0.082 & 0.147 & 0.277 \\
& M2KT\cite{14yang2023radiology} & 2023 & 0.224 & 0.179 & 0.151 & 0.371 & 0.078 & 0.153 & 0.261  \\
& DCL\cite{li2023DCL} & 2023 & 0.257 & 0.202 & 0.177 & 0.354 & 0.074 & 0.152 & 0.267  \\
& RGRG\cite{17tanida2023RGRG} & 2023 & 0.241 & 0.224 & 0.187 & 0.266 & 0.063 & 0.146 & 0.180 \\
& PromptMRG\cite{16jin2024promptmrg} & 2024 & \underline{0.275} & \underline{0.288} & \underline{0.246} & \underline{0.401} & \underline{0.098} & \textbf{0.160} & \underline{0.281} \\
\rowcolor{cyan!60!blue!20}
& \textbf{DDaTR(Ours)} & 2024 & \textbf{0.304} & \textbf{0.291} & \textbf{0.262(+0.016)} & \textbf{0.421} & \textbf{0.103} & \textbf{0.160} & \textbf{0.307} \\
\hline
\end{tabular}}
\vspace{-3mm}
\end{table*}

\subsection{Dynamic Feature Alignment Module}
\label{sec:DFAM}
Prior report typically contain valuable clinical prior knowledge that goes beyond what is directly observable in the image, including the clinician's focus and key diagnostic observations. Therefore, our goal is not merely to extract visual features from the prior image, but to enhance them by aligning with the semantic information in the report. Therefore, we propose DFAM, which enables the model to inject clinical semantics from the report into the pixel-level representation of the image, allowing the extracted prior features to better reflect clinical meaning. Our DFAM is inspired by the findings of the LAVT \cite{yang2022lavt}, which demonstrated that early fusion of image-text features within the intermediate layers of a vision encoder can more effectively align visual features with text. During the prior feature encoding process, we use the prior observation classification task as a supervisory signal. The structure of the DFAM is illustrated Fig. \ref{fig:DFAM}, which consists of two components: feature alignment and dynamic fusion. 

At the first step, the input features from the two modalities—visual features $F_{prior}^{m} \in \mathbb{R}^{ C \times H \times W}$ at $m^{th}$ stage and textual features $F_{txt} \in \mathbb{R}^{ C \times L}$—are aligned. Specifically, $F_{prior}^{m}$ is mapped as queries, and $F_{txt}$ is mapped as keys and values to conduct scaled dot-product attention \cite{2017Transformer}. This design choice is made to ensure that the fused representation remains primarily image-driven and facilitates the subsequent perception of visual difference. The textual features are incorporated to inject clinical semantic of reports into the visual representations, thereby promoting better alignment between the prior image features and the underlying clinical intent. We define $P$ as the operation of projection. This process is represented by the following equations: 
{\small
\begin{equation}
    Q = P^q(\text{Reshape}(F_{prior}^{m}));K = P^k(F_{txt});V = P^v(F_{txt})
  \label{eq_PF13}
\end{equation}
\vspace{-4mm}
\begin{equation}
    F_{att} = P^t(\text{Softmax}\left(\frac{Q^T K}{\sqrt{C}}\right) V^T)
  \label{eq_PF16}
\end{equation}}
where $P^k$ and $P^v$ are 1×1 convolution. $P^q$ and $P^t$ are the 1×1 convolution followed by instance normalization. $F_{att}$ represents the output features after this process. This output $F_{att} \in \mathbb{R}^{H·W \times C}$ has the same spatial resolution as the original visual features but integrates semantic information from the text at each image pixel. This design is aligned with our goal of injecting report-level semantic guidance into the visual representation, allowing each spatial position in the image to be enriched with contextually relevant textual semantics. Next, $F_{att}$ are further integrated with the feature mapped from original image features $F_{prior}^{m}$ as follows:
{\small
\begin{equation}
    F_{fa} = P^f (P^a(\text{Reshape}(F_{prior}^{m})) \odot F_{att})
  \label{eq_PF18}
\end{equation}}
where $P^a$ and $P^f$ are 1×1 convolution followed by ReLU, \( \odot \) denotes element-wise multiplication and $F_{fa} \in \mathbb{R}^{H·W \times C}$ represents the output of the first feature alignment part. 

Secondly, the $F_{fa}$ is fused with the original input image feature $F_{prior}^{m}$ using a gating mechanism.
Specifically, for the aligned features $F_{fa}$, the process has two blocks $B_1$ and $B_2$. The output features $F_{dy}$ are then combined with $F_{prior}^{m}$ through element-wise multiplication, ultimately producing the final output features $F_{align}^{m}$ of the DFAM at $m^{th}$ stage of prior encoder. This process is described as follows:
{\small
\begin{equation}
    F_{align}^{m} = B_2(B_1(F_{fa})) \odot F_{fa}
\end{equation}}
where $B_2(B_1(F_{fa}))$ denotes the output features of gating mechanism. $B_1$ is the operation of 1x1 convolution followed by ReLU. $B_2$ is the operation of 1x1 convolution with Tanh.

\subsection{Dynamic Difference-aware Module}
\label{sec:DDAM}

To more effectively capture the multi-level progress information between prior and current images, we propose the Dynamic Difference-aware Module (DDAM). Similar to the DFAM, the DDAM is also integrated at every stage of the visual encoder to enable perception at both low-level and high-level features. Furthermore, this module can adaptively enhance the texture details of both visual features. The structure of the DDAM is illustrated in Fig. \ref{fig:DDAM} and comprises two components: difference awareness and dynamic fusion.

In the first part, we employ the enhanced prior visual features $\hat{F}_{prior}^{m}$ processed using DFAM. Inspired by Fusionmamba \cite{xie2024fusionmamba}, the Learnable Descriptive Convolution (LDConv) \cite{huang2022LDConv} improves texture processing on input feature maps by incorporating learnable descriptors into vanilla convolution operations. Therefore, we also incorporate the LDConv in our DDAM to improve texture detail features. “the texture details” refer to fine-grained visual patterns in the images, such as edges, contours, and subtle structural variations. Specifically, given the current visual feature $F_{cur}^{m}$ and the prior enhanced visual feature $\hat{F}_{prior}^{m}$, we first apply two LDConv modules to enhance each feature individually, resulting in the enhanced features $F_{enc}$ and $F_{enp}$:
\begin{equation}
\small
    F_{enc} = LDC_c(F_{cur}^{m}) ; F_{enp} = LDC_p(\hat{F}_{prior}^{m})
\end{equation}
Next, we compute the pixel differences between the enhanced features $F_{enc}$ and $F_{enp}$:
\begin{equation}
\small
    F_{pd} = \text{Sigmoid}(\text{AvgPool}(F_{enc} - F_{enp}))
\end{equation}
where $F_{pd}$ is the output feature of pixel differences. Then, the difference feature $F_{pd}$ is applied to the two enhanced features $F_{enc}$ and $F_{enp}$ to amplify the distinctions between them, obtaining the current difference-aware feature $F_{enc} \odot F_{pd}$ and the prior one $F_{enp} \odot F_{pd}$. Finally, we concatenate them to obtain the difference-aware feature. The process is as follows:
\begin{equation}
\small
    F_{da} = P^d (Cat (F_{enc} \odot F_{pd}, F_{enp} \odot F_{pd}))
\end{equation}
where $Cat(\cdot)$ represents the operation of concatenation. $P^d$ is a 1×1 convolution followed by ReLU activation.

Secondly, the dynamic fusion part of the DDAM is the same as DFAM, designed to ensure that the difference-aware features do not overwhelm the current visual features while allowing a controlled flow of longitudinal information into the next stage. This module dynamically integrates prior information, allowing the model to adaptively select the most relevant features. Specifically, $F_{cur}^{m}$ represents the visual features, $F_{da}$ means the
difference-aware visual features including longitudinal features. When we fuse the $F_{cur}$ and $F_{da}$, we dynamically add $F_{da}$ to $F_{cur}$ using a gating mechanism. $f_{dy}$ represents our gating module, and the fusion output is $F_{dy} = f_{dy}(F_{da})$, where $F_{dy}$ represents the output weights generated by our gating module. Then, using ‘$\alpha$’ indicates whether the input has a prior examination; it is set to 1 if a prior exists, and 0 otherwise. So the fused process is as follows: 
{\small
\begin{equation}
    \hat{F}_{cur} = F_{cur} + \alpha(F_{dy} \odot F_{da}),
\end{equation}
}
where $\hat{F}_{cur}$ means the final visual feature after the dynamic fusing process. We can see that $F_{dy}$ can adjust the prior features when the prior examination is available. 

\section{Experiments}

\subsection{Datasets and Metrics}

\textbf{MIMIC-CXR}. The MIMIC-CXR dataset \cite{27johnson2019mimic} consists of 473, 057 images and 227, 835 reports from 63, 478 patients. We utilize the official data split (270,790 for training, 2,130 for validation, and 3,858 for testing). To construct the longitudinal data, we obtain the prior image and report by order different cases according to corresponding timestamps. For each study, we select the most recent historical study as the prior, including consideration of lateral views as well. As a result, the training set contains 146,061 studies with prior exams and 124,729 without; the validation set includes 1,181 with prior exams and 949 without; and the test set consists of 3,153 with prior exams and 705 without.

\textbf{Longitudinal-MIMIC}. This dataset \cite{zhu2023Lr2gen} only contains Longitudinal cases. The official data is split as follows: the training set includes 26,156 patients and 92,374 samples, the validation set comprises 203 patients and 737 samples, and the test set contains 266 patients and 2,058 samples.

\textbf{IU-Xray}. The dataset\cite{iuxray} contains a total of 2,955 samples, all are without longitudinal exam. Following the construction strategy of PromptMRG, we reorganized the dataset into 4,168 image and report pairs, all of which are used as the test set.

\textbf{Evaluation metrics}. We employed the same Natural Language Generation (NLG) metrics as those used in previous studies: BLEU scores \cite{papineni2002bleu}, METEOR \cite{banerjee2005meteor}, ROUGE-L \cite{lin2004rouge}. These metrics were evaluated following the standard evaluation protocol\footnote{\url{github.com/tylin/coco-caption}.}. We also assess clinical efficacy (CE) of reports utilizing the CheXbert \cite{smit2020CheXbert}. CheXbert can generate disease labels and attributes for each report. Following previous work \cite{25hou2023recap}, we map `positive' and `uncertain' to the positive label and map `negative' and `blank' to the negative label. Then, based on the report classification results, we can calculate the precision (P), recall (R), and F1-score (F1).

\textbf{Experiment Settings}. Our code, based on the Pytorch framework \cite{paszke2019pytorch}, was trained using a single NVIDIA RTX A40 GPU. For all methods used in our comparison, we strictly followed the default configurations reported in their respective original papers without any modifications. To ensure fair and consistent evaluation, we adopted the same training settings as the baseline PromptMRG across all our experiments, including those conducted on different datasets and with various visual backbones. Specifically, all input images were resized to 256 × 256, followed by data augmentations such as 224 × 224 random cropping and random rotation within ±5 degrees. The maximum input text length was set to 100 words. We employed the AdamW optimizer \cite{loshchilov2017adamw} with a weight decay of 0.05. The initial learning rate was set to 5e-5, and all models were trained for 10 epochs with a batch size of 16. Overall loss function combines the Language Modeling (LM) loss $\mathcal{L}_{LM}$ \cite{blip} for generation and standard Cross-Entropy (CE) loss $\mathcal{L}_{CE}$ for current and prior disease classification. The total loss $\mathcal{L}$ is defined as follows:

{\small
\begin{equation}
    \mathcal{L} = \mathcal{L}_{LM}+ w(\mathcal{L}_{CE}^{cur} + \mathcal{L}_{CE}^{prior}),
\end{equation}
}

where $w$ is set to 4, consistent with the setting used in PromptMRG.

\begin{table}[t]
\setlength{\tabcolsep}{2pt}
\centering
\caption{Performance Comparison on three different visual backbones: ResNet101, SwinT-B, and EfficientNetv2. Vanilla represents that visual features were extracted using original pretrained models, coupled with the frozen pretrained BERT for text feature extraction, with subsequent concatenation of these features.}
\label{tab:vis com}
\resizebox{0.49\textwidth}{!}{
\begin{tabular}{c|c|c|ccc|ccc}
\hline
\multirow{2}{*}{Datasets} & \multirow{2}{*}{\makecell[c]{Visual\\Backbone}} & \multirow{2}{*}{Methods} & \multicolumn{3}{c|}{Clinical Efficacy} & \multicolumn{3}{c}{Natural Language Generation} \\
\cline{4-9}
& &  & Precision & Recall & F1 & BLEU\_4 & METEOR & ROUGE\_L \\ \hline
\multirow{6}{*}{\makecell[c]{MIMIC\\-CXR}} & \multirow{2}{*}{ResNet101} & Vanilla & 0.395 & 0.426 & 0.389 & 0.108 & 0.156 & 0.267 \\
& & Ours & \textbf{0.411} & \textbf{0.449} & \textbf{0.421 (+0.032)} & \textbf{0.109} & \textbf{0.158} & \textbf{0.269} \\ \cline{2-9}
& \multirow{2}{*}{Efficientv2} & Vanilla & 0.384 & 0.430 & 0.396 &  0.108 & 0.157 & 0.267 \\ 
& & Ours & \textbf{0.440} & \textbf{0.451} & \textbf{0.433 (+0.037)} & \textbf{0.111} & \textbf{0.160} & \textbf{0.270} \\ \cline{2-9}
& \multirow{2}{*}{SwinT-B} & Vanilla & 0.400 & 0.417 & 0.392 & 0.109 & 0.157 & 0.269 \\
& & Ours & \textbf{0.438} & \textbf{0.465} & \textbf{0.441 (+0.049)} & \textbf{0.113} & \textbf{0.162} & \textbf{0.272} \\ \hline
\multirow{6}{*}{\makecell[c]{Longitudinal\\-MIMIC}} & \multirow{2}{*}{ResNet101} & Vanilla & 0.487 & 0.544 & 0.486 & 0.101 & 0.151 & 0.261 \\
& & Ours & \textbf{0.519} & \textbf{0.577} & \textbf{0.516 (+0.030)} & \textbf{0.105} & \textbf{0.155} & \textbf{0.266} \\ \cline{2-9}
& \multirow{2}{*}{Efficientv2} & Vanilla & 0.484 & 0.552 & 0.487 & 0.101 & 0.151 & 0.264 \\ 
& & Ours & \textbf{0.491} & \textbf{0.568} & \textbf{0.503 (+0.016)} & \textbf{0.106} & \textbf{0.154} & \textbf{0.268} \\ \cline{2-9}
& \multirow{2}{*}{SwinT-B} & Vanilla & 0.506 & 0.563 & 0.504 & 0.102 & 0.155 & 0.263 \\
& & Ours & \textbf{0.539} & \textbf{0.584} & \textbf{0.527 (+0.023)} & \textbf{0.105} & \textbf{0.156} & \textbf{0.266}\\ 
\hline
\end{tabular}}
\vspace{-5mm}
\end{table}

\begin{table}[t]
\setlength{\tabcolsep}{2.5pt}
\centering
\caption{Ablation Study. w/o MSF indicates without multi-stage fusion. w/o DAM indicates without difference-aware module. w/o FAM indicates without feature alignment module. w/o DF indicates without dynamic fusion of both DFAM and DDAM.} 
\label{tab: ablation study}
\resizebox{0.49\textwidth}{!}{
\begin{tabular}{c|c|ccc|cccc}
\hline
\multirow{2}{*}{SwinT-B} & \multirow{2}{*}{Model} & \multicolumn{3}{c|}{Clinical Efficacy} & \multicolumn{4}{c}{Natural Language Generation}\\
\cline{3-5}\cline{6-9}
 &  & \multicolumn{1}{c}{Precision} & \multicolumn{1}{c}{Recall} & \multicolumn{1}{c|}{F1} & \multicolumn{1}{c}{BLEU-1} & \multicolumn{1}{c}{BLEU-4} & \multicolumn{1}{c}{METEOR} & \multicolumn{1}{c}{ROUGE-L} \\
\hline
\multirow{5}{*}{\makecell[c]{MIMIC\\-CXR}} & w/o MSF & 0.415 & 0.419 & 0.408 & 0.395 & 0.107 & 0.154 & 0.267 \\
& w/o DAM & 0.413 & 0.417 & 0.402 & 0.381 & 0.100 & 0.150 & 0.258 \\
& w/o FAM & 0.409 & 0.455 & 0.416 & 0.400 & 0.111 & 0.157 & 0.270  \\
& w/o DF & 0.402 & 0.465 & 0.421 & 0.396 & 0.109 & 0.158 & 0.269 \\
& DDaTR & \textbf{0.438} & \textbf{0.465} & \textbf{0.441} & \textbf{0.401} & \textbf{0.113} & \textbf{0.162} & \textbf{0.272} \\
\hline
\multirow{5}{*}{\makecell[c]{Longitudinal\\-MIMIC}} & w/o MSF & 0.535 & 0.556 & 0.515 & 0.390 & 0.103 & 0.155 & 0.265 \\
& w/o DAM & 0.527 & 0.556 & 0.512 & 0.388 & 0.101 & 0.153 & 0.262 \\
& w/o FAM & 0.531 & 0.567 & 0.518 & 0.386 & 0.103 & 0.156 & 0.265  \\
& w/o DF & 0.522 & 0.583 & 0.521 & 0.389 & \textbf{0.106} & 0.156 & 0.266 \\
& DDaTR &  \textbf{0.539} & \textbf{0.584} & \textbf{0.527} & \textbf{0.396} & 0.105 & \textbf{0.156} & \textbf{0.266} \\
\hline
\end{tabular}}
\vspace{-3mm}
\end{table}

\subsection{Experimental Results}

\textbf{Comparison on both RRG (MIMIC-CXR, IU-XRay) and LRRG (Longitudinal-MIMIC) benchmarks.} We compare our method with a wide range of classical and SOTA methods on MIMIC-CXR, IU-XRay and Longitudinal-MIMIC, and report their results in Tab. \ref{tab:main result}. 
For the MIMIC-CXR dataset, we follow the evaluation established by RECAP, calculating the Macro score for the CE metrics. For the IU-Xray dataset, we use the same setting as PromptMRG that did not conduct any training or fine-tuning on it, but directly tested our model which trained on MIMIC-CXR. For the Longitudinal-MIMIC dataset, our approach follows the evaluation methodology used in LR2Gen, where the Micro score is calculated for the CE metrics. By comparison, our method achieves SOTA on CE metrics, significantly outperforming other approaches on both RRG and LRRG. We performed a statistical significance analysis on the MIMIC-CXR test set by calculating the p-value of the F1 scores across all longitudinal cases. The result is $P_{F1}$=0.0415 \textless 0.05 comparing with previous SOTA promptmrg, indicating that the performance improvement of our method is statistically significant. This not only demonstrates that utilizing longitudinal information for report generation tasks yields reports that more closely align with the groundtruth, but also highlights our method's capability to more accurately capture the diagnostic intent of the radiologists. For example, our method with SwinT-B on MIMIC-CXR achieves 0.441 of F1\_macro CE score with an absolute improvement of 0.048. Our method with SwinT-B on Longitudinal-MIMIC achieves 0.527 of F1\_micro CE score with an absolute improvement of 0.035. 

\begin{table}[t]
    \centering
    \setlength{\tabcolsep}{0.8pt}
    \scriptsize
    \caption{Performance Metrics on MIMIC-CXR Dataset. Comparison with Large language models.}
    \label{tab:LLM}
    \resizebox{0.49\textwidth}{!}{
    \begin{tabular}{ccccccc}
    \hline
    \textbf{Model} & BLEU-1$\uparrow$ & BLEU-4$\uparrow$ & ROUGE-L$\uparrow$ & METEOR$\uparrow$ & F1-RadGraph$\uparrow$ & RadCliQ$\downarrow$ \\
    \hline
    RadFM\cite{wu2023RadFM} & 0.221 & 0.056 & 0.205 & 0.204 & 0.182 & 1.687 \\
    LLaVA-Med\cite{li2024llavamed} & 0.193 & 0.010 & 0.139 & 0.132 & 0.063 & 2.155 \\
    Med-Flamingo\cite{moor2023medflamingo} & 0.224 & 0.019 & 0.145 & 0.140 & 0.071 & 2.116 \\
    MedDr\cite{he2024meddr} & 0.322 & 0.072 & 0.226 & \textbf{0.238} & \textbf{0.224} & 1.538 \\
    \hline
    \textbf{DDaTR(Ours)} & \textbf{0.401} & \textbf{0.113} & \textbf{0.272} & 0.162 & 0.198 & \textbf{1.141} \\
    \hline
    \end{tabular}}
    \vspace{-3mm}
\end{table}

\begin{table}[t]
    \centering
    \setlength{\tabcolsep}{0.8pt}
    \scriptsize
    \caption{Performance on ReXrank Dataset. Comparison with existing specialist RRG Models.}
    \label{tab: rexrank}
    \resizebox{0.49\textwidth}{!}{
    \begin{tabular}{cccccccc}
    \hline
    \textbf{Model} & Year & BLEU-2$\uparrow$ & F1-RadGraph$\uparrow$ & BertScore$\uparrow$ & SembScore$\uparrow$ & 1/RadCliQ-V1$\uparrow$ & GREEN$\uparrow$ \\
    \hline
    RadFM\cite{wu2023RadFM} & 2023 & 0.087 & 0.109 & 0.313 & 0.259 & 0.650 & 0.185 \\
    VLCI\cite{vlci} & 2023 & 0.136 & 0.140 & 0.304 & 0.305 & 0.680 & 0.256 \\
    CVT2Dis.\cite{CVT2Dis} & 2023 & 0.126 & 0.149 & 0.331 & 0.329 & 0.719 & 0.268 \\
    RaDialog\cite{radialog} & 2023 & 0.127 & 0.172 & \underline{0.363} & 0.387 & 0.799 & 0.273 \\
    LLM-CXR\cite{llmcxr} & 2024 & 0.037 & 0.046 & 0.181 & 0.156 & 0.516 & 0.043 \\
    MAIRA-2\cite{maira2} & 2024 & 0.088 & 0.131 & 0.308 & 0.339 & 0.694 & 0.224 \\
    CheXagent\cite{chexagent} & 2024 & 0.113 & 0.148 & 0.346 & 0.347 & 0.741 & 0.257 \\
    MedVersa\cite{medversa} & 2024 & \underline{0.209} & \textbf{0.273} & \textbf{0.448} & \textbf{0.466} & \textbf{1.103} & \textbf{0.374} \\
    \hline
    \textbf{DDaTR(Ours)} & 2024 & \textbf{0.210} & \underline{0.198} & 0.362 & \underline{0.465} & \underline{0.874} & \underline{0.293} \\
    \hline
    
    \end{tabular}}
    \vspace{-3mm}
\end{table}

\begin{figure*}[t]
    \centering
    \includegraphics[scale=.48]{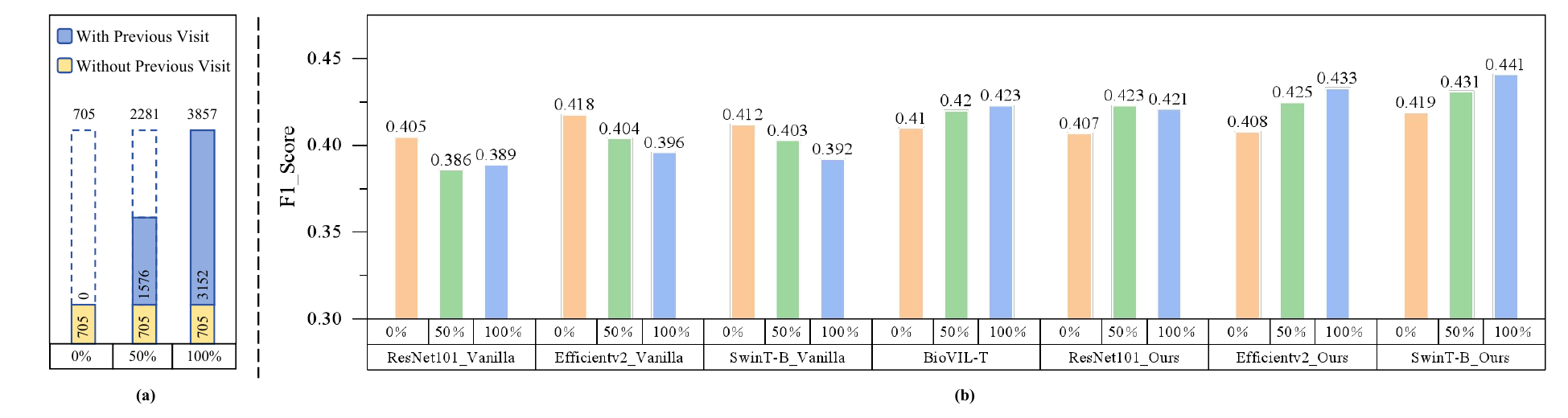}
    \vspace{-3mm}
    \caption{(a) Illustrates different test settings. Each column represents a specific test setting, with n\% indicating the proportion of longitudinal reports used. The total number of samples for each setting is shown at the top of each column. (b) Performance Comparison on different test settings.}
    \label{fig:barchart}
    \vspace{-5.5mm}
\end{figure*}

In the task of RRG, CE metrics are considered more important than NLG metrics. This is because different radiologists exhibit variations in language expression, terminology, and writing styles when composing reports. NLG scores mainly measure the similarity between texts at the sentence level, and minor improvements in NLG scores do not necessarily reflect a more accurate diagnosed of report. However, the precision in describing key observations in report is crucial. CE metrics specifically measure the diagnostic accuracy of reports concerning corresponding attributes. This has been overlooked in some studies which focused on improving NLG scores instead of CE metrics. Therefore, although the NLG scores of our method on the MIMIC-CXR dataset achieved performance just comparable to SOTA methods without showing a significant improvement, excelling in CE metrics could demonstrate our model's superiority and practical utility.

We analyze this as follows: a review of existing methods reveals that higher NLG scores often result from enhancements in the decoder. Thus, under similar data processing and training strategies, the NLG scores largely depend on the ability of the model's decoder, which includes aspects such as the architecture, the tokenizer, and the pre-trained models. Our research primarily focuses on how to encode longitudinal information effectively, aligning input features more closely with medical intent. Consequently, we did not make specific structural modifications to the decoder but instead followed the decoder used in PromptMRG. Furthermore, our zero-shot results on IU X-Ray also outperform previous SOTA, which does not contain longitudinal information. This highlights the robustness and generalizability of our approach and further demonstrates that our model not only substantially improve longitudinal report generation, but also maintain the advanced performance in single-report scenarios.

\textbf{Comparison on different visual backbones}. Our proposed algorithmic framework is adaptable to all multi-scale visual feature extraction networks. To validate its effectiveness, we conducted comparative experiments using three prominent visual backbones: ResNet101 \cite{he2016deep}, SwinT-B \cite{liu2021swin}, and EfficientNetv2 \cite{tan2021efficientnetv2}. These networks were selected because they are commonly employed for visual feature extraction in prior RRG research. We compared the performance outcomes between using standard vanilla visual networks and our method under the same baseline framework. In the vanilla approach, visual features were extracted using the original network architectures (e.g., ResNet101) and also fine-tuned during training using the same strategy as ours. For textual features, a pretrained BERT model was also used in a frozen state, and the visual and textual features were subsequently concatenated for downstream tasks. Experimental results, presented in Tab. \ref{tab:vis com}, demonstrate that our method outperforms the vanilla encoder across all three backbones by a significant margin.


\textbf{Comparison on ReXrank}. To compare our method with the latest approaches in RRG, we evaluated DDaTR on the official test set of MIMIC-CXR provided by the ReXrank leaderboard \cite{rexrank}, which including the comprehensive metrics, such as BERTScore\cite{bertscore}, SembScore\cite{smit2020CheXbert}, and GREEN\cite{green}. The results are presented in Tab. \ref{tab: rexrank}. Our model ranks second on the leaderboard. While MedVersa \cite{medversa} significantly outperforms all other methods on the leaderboard, it is trained on 10 public datasets and built upon LLM. Due to the substantial differences in both data amount and model size, direct comparison is not entirely fair. Compared to other task-specific LLMs that are explicitly trained on RRG data—such as LLM-CXR\cite{llmcxr}, CheXagent\cite{chexagent}, RaDialog\cite{radialog} and MAIRA-2\cite{maira2}—our method achieves consistently superior performance across all evaluation metrics, further demonstrating the effectiveness of our approach under a more lightweight setting. 

\textbf{Comparison on complexity}. To compare model complexity, we report the trainable parameters for ours and PromptMRG as follows: PromptMRG (220M), Ours with Swin-T (360M), Ours with EfficientNetV2 (247M), and Ours with ResNet101 (419M). 
Notably, when using EfficientNetV2, our model maintains a comparable parameter count (247M) to PromptMRG (220M), yet still achieves significantly better performance as shown in Tab. \ref{tab:vis com}. This indicates that the performance gains of our method are not merely attributed to increased model complexity. In fact, in Tab. \ref{tab:vis com} Longitudinal-MIMIC, ours with ResNet101 has the highest number of parameters but does not perform as well as the other backbones, further emphasizing that simply increasing model complexity does not guarantee better longitudinal modeling performance. Furthermore, PromptMRG is designed for single-period input, while our method handles longitudinal data, which is acceptable to introduce some additional parameters.

\begin{figure*}[t]
    \centering
    \includegraphics[width=1\textwidth]{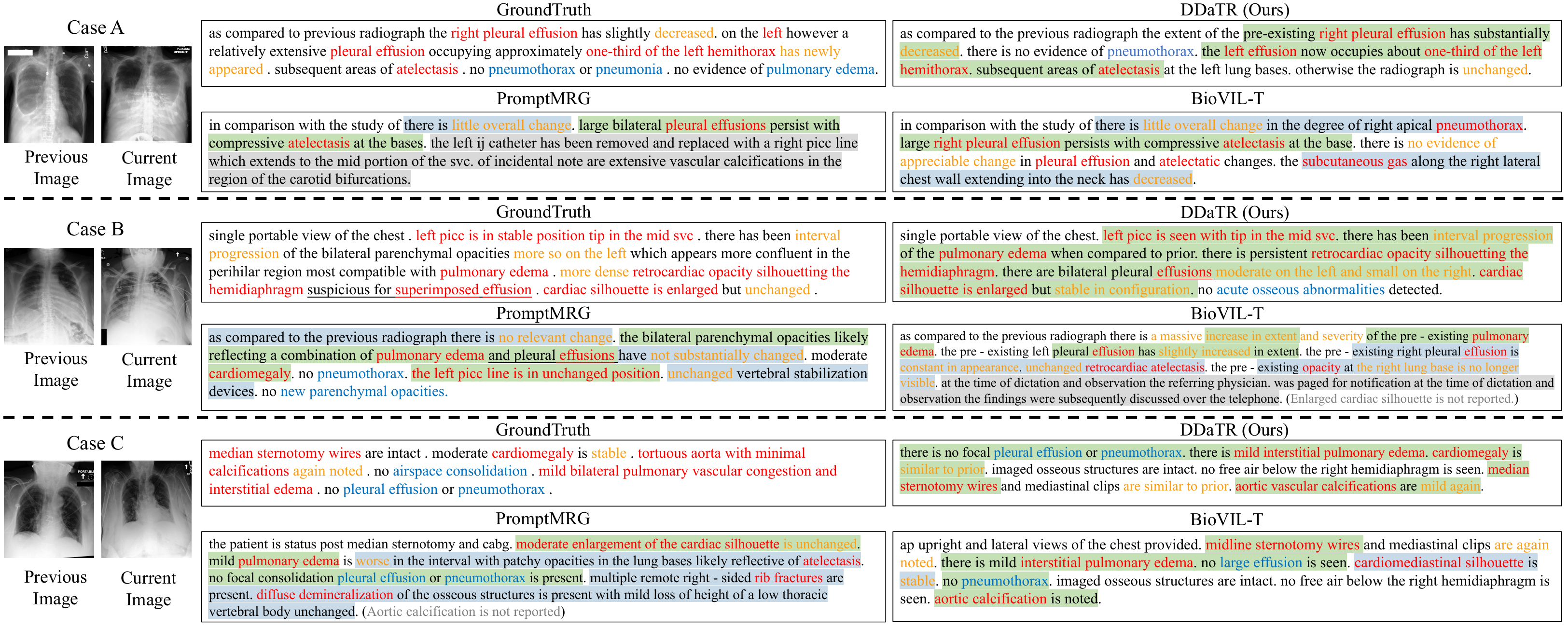}
    \caption{Case studies comparing radiology report predictions generated by our proposed DDaTR and two baseline methods. \textcolor{bluetext}{\textbf{Blue}} texts indicates negative observations, the \textcolor{redtext}{\textbf{red}} texts indicates positive observations, and the \textcolor{yellowtext}{\textbf{yellow}} texts denotes the progressions. The \sethlcolor{bggreen}\hl{green background} indicates correctly predicted sentences, the \sethlcolor{bgblue}\hl{blue background} denotes incorrectly predicted sentences, and the \sethlcolor{bggray}\hl{gray background} highlights overstatements.}
    \label{fig:casestudy}
    \vspace{-5.5mm}
\end{figure*}

\textbf{Comparison on different proportions of longitudinal cases}. We conducted assessments not only on standard MIMIC-CXR benchmark but also under various settings with different proportions of longitudinal cases. As shown in Fig. \ref{fig:barchart} (a), we split the testset of MIMIC-CXR. 0\% indicates that the test data without longitudinal information. Conversely, 100\% means all test cases. At 50\%, the test set is composed by randomly selecting half of the reports with prior exams.
In the above three testing settings, we evaluated the performance of different encoding methods within the baseline frameworks. Specifically, we tested the effectiveness of three prominent visual backbones using both the standard vanilla approach and our method, alongside another pretrained encoder named BioVIL-T \cite{bannur2023Biovil}, which is designed for LRRG specifically. It is important to note that BioVIL-T utilizes image inputs of 448 x 448 pixels. Our method and other mainstream approaches use a standard input size of 224 x 224 pixels. The larger the image resolution, the richer the image details captured, which naturally provides an additional advantage to BioVIL-T. As illustrated in Fig. \ref{fig:barchart} (b), despite BioVIL-T's use of larger images, our method still surpasses BioVIL-T. Additionally, as shown in Fig. \ref{fig:barchart} (b), the vanilla methods underperform in generating multi-period reports. In contrast, our method not only enhances accuracy in single-period report generation but also significantly improves multi-period report generation.

\textbf{Comparison with medical large language models}. Currently, the field of large language models is rapidly evolving, leading to the emergence of numerous substantial medical models aimed at solving a variety of medical image-text tasks. To more comprehensively assess the value of our method, we conducted a comparison with existing medical large language model approaches, including RadFM \cite{wu2023RadFM}, LLaVA-Med \cite{li2024llavamed}, Med-Flamingo \cite{moor2023medflamingo} and MedDr \cite{he2024meddr}. The evaluation metrics follow those used in MedDr. As indicated in Tab. \ref{tab:LLM}, in the RRG task, our specialized model demonstrates a distinct advantage over these larger models, particularly in terms of the RadCliQ score \cite{yu2023RadCliQ} (a comprehensive metric that integrates multiple metrics). This suggests that while large models possess robust generalizability and the capability to handle multiple medical image-text tasks, their performance in specific tasks is still inferior to that of dedicated models. Additionally, large models need longer training and inference times, as well as more hardware requirements. In practical applications, particularly for RRG tasks, large models currently cannot fully replace task-specific models.

\subsection{Ablation Study and Case Study}

\textbf{Effect of multi-stage fusion}. We initially validated the method utilizing multi-stage fusion. Our analysis compared two approaches: one that aligns multimodal features and detects differences only at the final visual stage, and another that applies feature alignment and difference detection across all visual stages. This comparison helps highlight the difference between our early fusion strategy that integrated within the visual encoder and the late fusion approach adopted by previous LRRG methods. As shown in Tab. \ref{tab: ablation study}, compared to single-stage feature integration, multi-stage fusion achieved an increase in both the NLG and CE metrics (0.033$\uparrow$  on F1 of MIMIC-CXR and 0.012$\uparrow$ on F1 of Longitudinal-MIMIC), highlighting the importance of leveraging multi-level visual features for enhancing longitudinal representation learning.

\begin{figure*}[t]
    \centering
    \includegraphics[width=0.94\textwidth]{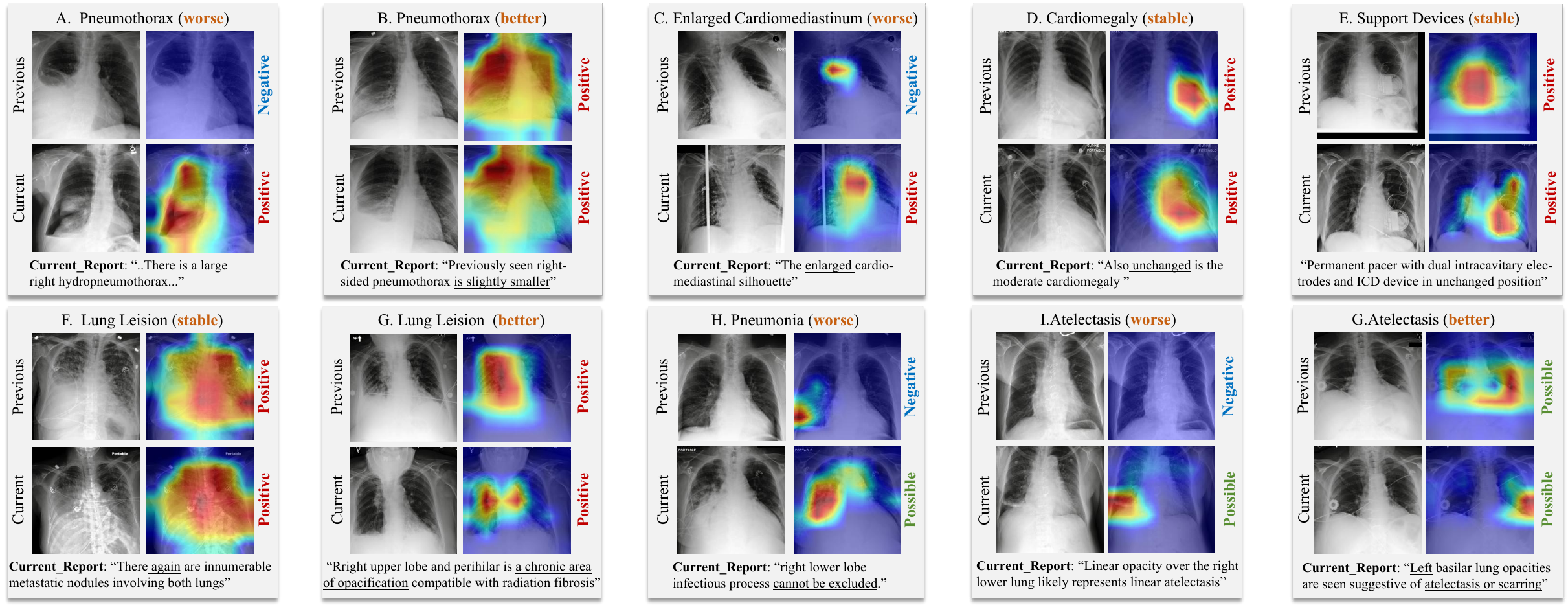}
    \vspace{-3mm}
    \caption{Grad-CAM visualization examples. Each sub-figure is titled with the case ID, diagnostic class, and progression. The top row displays the previous image and its Grad-CAM map; the bottom row shows the current ones. Diagnostic results (positive, negative, or possible) are provided on the right. At the bottom, the section of current report from the physician are provided for reference, with key information highlighted using `\_'.}
    \label{fig:visualization}
    \vspace{-5mm}
\end{figure*}

\textbf{Effect of DAM and FAM}. DAM and FAM represent the configurations where dynamic fusion is removed from the DDAM and DFAM modules, respectively. In our ablation studies, we separately removed DAM and FAM to assess their impacts on the model's performance. As indicated in Tab. \ref{tab: ablation study}, the FAM led to an F1 increase of 0.025 on the MIMIC-CXR and 0.011 on the Longitudinal-MIMIC. These results demonstrate the effectiveness of FAM in extracting prior visual features that are better aligned with clinical semantics. Similarly, DAM resulted in an an increase of 0.039 in the F1 score on the MIMIC-CXR and 0.018 on the Longitudinal-MIMIC, demonstrating its effectiveness in enhancing the model’s ability to capture temporal changes. The comparison between the two experiments indicates that the DAM plays a more critical role, with substantial impact on both the CE and NLG metrics. Moreover, the FAM, which introduces clinical intent of the prior report, shows a notable improvement in CE but has limited effect on NLG. This suggests that the prior report primarily contributes to enhancing diagnostic accuracy.

\textbf{Effect of dynamic fusion}. We evaluated the effect of the dynamic fusion mechanism itself, which adaptively integrates features. As shown in Tab. \ref{tab: ablation study}, the overall improvement in F1 is primarily attributed to an increase in precision, while the recall remains unchanged. We attribute this to the effectiveness of the dynamic fusion mechanism that enables the model to better handle scenarios where prior information may be missing or noisy, thereby improving diagnostic precision.

\textbf{Case study}. Fig.\ref{fig:casestudy} presents a comparison of generated results. As can be observed, our method accurately captures the changes between two period images and provides precise descriptions. Notably, in case A, it can clearly articulate detailed descriptions, such as `one-third of the left hemithorax'. We analyzed the training data and found that around 10 cases contain this type of description, such as “the effusion occupies approximately one-third of the left hemithorax”, “occupying approximately one-third of the left hemithorax”. Although the occurrence is relatively rare, our method appears to have captured the associated patterns and is able to generate similar expressions during testing when encountering similar cases. In Case B, we observed that the generated reports still exhibit misalignment issues. Although ours is more accurate than those of the baselines, the ground truth indicates uncertainty regarding effusions, while all models including ours make a definitive statement. This highlights an ongoing challenge in distinguishing between definite and probable diagnoses, which can vary due to differing interpretations of uncertainty among radiologists with different levels of experience. In Case C, we observed that the generated reports still exhibit hallucinations issues. The model occasionally generates additional negative findings or extra information, such as "mediastinal clips". In addition, there are still many cases with diagnostic errors reflecting omissions or false positive issues.

\textbf{Visualization}. To better demonstrate how the model captures and utilizes longitudinal information, we conducted the visualization experiments of our DDATR. Specifically, we generated Grad-CAM \cite{gradcam} visualizations for both the current and previous images according to diagnostic labels, as shown in Fig. \ref{fig:visualization}. We present ten examples, analyzing different diagnostic categories and progressions. 

From these examples, we observe that in Case A and Case I, the model primarily focuses on the current image abnormalities for decision-making, without attending to features from the previous negative exam (less contribution). In contrast, in Case C and Case H, although the previous images also appeared negative for target diseases, the model shows slight attention to the corresponding regions in the previous images when diagnosing current abnormalities. This suggest that the model may tend to reference previous features when making decisions about enlarged (Case C) or when facing uncertainty (Case H). For the remaining examples, where both the previous and current images show positive findings, the model exhibits significant attention to abnormalities across both time points. These observations are consistent with the intended effect of our dynamic feature fusion mechanism.

Also, further analysis of cases where both the previous and current images are positive demonstrate our model’s ability to perceive temporal differences between the two longitudinal examinations by using DDAM. For example, in Case B, the visualization shows the slight reduction in right-sided pneumothorax compared to the previous image. In Case C, the results highlights an enlargement of cardiomediastinum.

Moreover, in Case G, the model first focuses on the bilateral lung bases in the previous image, accurately aligning with the input prior report statement: “Patchy bibasilar airspace opacities most likely reflect atelectasis.” In the current image, the model’s attention shifts to the left lung base, which aligns precisely with the corresponding detail in the current report mentioning “left basilar lung.” This strong alignment between generated and reference reports highlights the effectiveness of incorporating prior information (refined, reminded, and guided via DFAM and DDAM) as both informative priors and cross-modal alignment templates, enabling the model to better interpret current findings in context.

\section{Conclusion}

In this paper, we design a novel dynamic difference-aware temporal residual network (DDaTR) to better capture spatio-temporal correlations among longitudinal information. For temporal perception, our entire network dynamically integrates temporal information based on different prior image features. The residual connections are used to unidirectionally transmit temporal information. For spatial perception, we introduce two modules at each stage of the visual encoder. DFAM employs the entire prior report's features to guide the prior image encoding at each stage. The DDAM leverages differences between two image features to adaptively enhance the source image feature. To better evaluate the comprehensive performance of our network on both single-period and multi-period report generation tasks, we conducted assessments not only on the standard MIMIC-CXR benchmark but also under various settings with different proportions of longitudinal reports. This new evaluation demonstrates the model's versatility in both single-period and multi-period radiology report generation tasks. Extensive experiments and analyses indicate that our method aligns more closely with clinical applications and achieves SOTA results in both RRG and LRRG tasks.

However, our model still has certain limitations that require further investigation and improvement. First, the diagnostic accuracy of current report generation models including ours still does not reach the level of a real radiologist. Issues such as omissions, hallucinations, and misalignments still persist in the generated reports. Therefore, in clinical practice, such models are primarily used to assist radiologists in improving efficiency, with the generated reports still requiring manual review and correction. Secondly, although our method has comprehensive experiments on the X-ray report generation task, its effectiveness on other imaging modalities, such as ultrasound or CT, remains to be further explored and analyzed. Third, each component of our longitudinal framework still holds potential for further optimization, which could lead to further performance gains in future work.

\appendices

\bibliographystyle{IEEEtran}
\bibliography{ref}

\begin{thebibliography}{10}
\providecommand{\url}[1]{#1}
\csname url@samestyle\endcsname
\providecommand{\newblock}{\relax}
\providecommand{\bibinfo}[2]{#2}
\providecommand{\BIBentrySTDinterwordspacing}{\spaceskip=0pt\relax}
\providecommand{\BIBentryALTinterwordstretchfactor}{4}
\providecommand{\BIBentryALTinterwordspacing}{\spaceskip=\fontdimen2\font plus
\BIBentryALTinterwordstretchfactor\fontdimen3\font minus \fontdimen4\font\relax}
\providecommand{\BIBforeignlanguage}[2]{{%
\expandafter\ifx\csname l@#1\endcsname\relax
\typeout{** WARNING: IEEEtran.bst: No hyphenation pattern has been}%
\typeout{** loaded for the language `#1'. Using the pattern for}%
\typeout{** the default language instead.}%
\else
\language=\csname l@#1\endcsname
\fi
#2}}
\providecommand{\BIBdecl}{\relax}
\BIBdecl

\bibitem{esteva2019guide}
A.~Esteva, A.~Robicquet, B.~Ramsundar \emph{et~al.}, ``A guide to deep learning in healthcare,'' \emph{Nature medicine}, vol.~25, no.~1, pp. 24--29, 2019.

\bibitem{sloan2024automated}
P.~Sloan \emph{et~al.}, ``Automated radiology report generation: A review of recent advances,'' \emph{IEEE Reviews in Biomedical Engineering}, 2024.

\bibitem{zhu2023Lr2gen}
Q.~Zhu, T.~S. Mathai \emph{et~al.}, ``Utilizing longitudinal chest x-rays and reports to pre-fill radiology reports,'' in \emph{MICCAI}, 2023, pp. 189--198.

\bibitem{dalla2023controllable}
F.~Dalla~Serra, C.~Wang, F.~Deligianni \emph{et~al.}, ``Controllable chest x-ray report generation from longitudinal representations,'' in \emph{EMNLP}, 2023.

\bibitem{wang2024hergen}
F.~Wang, S.~Du, and L.~Yu, ``Hergen: Elevating radiology report generation with longitudinal data,'' \emph{ECCV}, 2024.

\bibitem{bannur2023Biovil}
S.~Bannur \emph{et~al.}, ``Learning to exploit temporal structure for biomedical vision-language processing,'' in \emph{CVPR}, 2023, pp. 15\,016--15\,027.

\bibitem{24nicolson2023longitudinal}
A.~Nicolson, J.~Dowling, D.~Anderson, and B.~Koopman, ``Longitudinal data and a semantic similarity reward for chest x-ray report generation,'' \emph{Informatics in Medicine Unlocked}, vol.~50, p. 101585, 2024.

\bibitem{25hou2023recap}
W.~Hou, Y.~Cheng, K.~Xu, W.~Li, and J.~Liu, ``Recap: Towards precise radiology report generation via dynamic disease progression reasoning,'' in \emph{EMNLP}, 2023, pp. 2134--2147.

\bibitem{06plummer2015flickr30k}
B.~A. Plummer, L.~Wang, C.~M. Cervantes, J.~C. Caicedo \emph{et~al.}, ``Flickr30k entities: Collecting region-to-phrase correspondences for richer image-to-sentence models,'' in \emph{CVPR}, 2015, pp. 2641--2649.

\bibitem{07vinyals2015show}
O.~Vinyals, A.~Toshev, S.~Bengio, and D.~Erhan, ``Show and tell: A neural image caption generator,'' in \emph{CVPR}, 2015, pp. 3156--3164.

\bibitem{11chen2020R2gen}
Z.~Chen, Y.~Song, T.~Chang, and X.~Wan, ``Generating radiology reports via memory-driven transformer,'' in \emph{EMNLP}, 2020, pp. 1439--1449.

\bibitem{chen2021R2GenCMN}
Z.~Chen, Y.~Shen, Y.~Song, and X.~Wan, ``Cross-modal memory networks for radiology report generation,'' in \emph{ACL}, 2021, pp. 5904--5914.

\bibitem{qin2022CMMRL}
H.~Qin and Y.~Song, ``Reinforced cross-modal alignment for radiology report generation,'' in \emph{Findings of the ACL}, 2022, pp. 448--458.

\bibitem{wang2023metransformer}
Z.~Wang, L.~Liu, L.~Wang, and L.~Zhou, ``Metransformer: Radiology report generation by transformer with multiple learnable expert tokens,'' in \emph{CVPR}, 2023, pp. 11\,558--11\,567.

\bibitem{20liu2021auto}
F.~Liu, C.~You \emph{et~al.}, ``Auto-encoding knowledge graph for unsupervised medical report generation,'' \emph{NeurIPS}, vol.~34, pp. 16\,266--16\,279, 2021.

\bibitem{22yang2022knowledge}
S.~Yang, X.~Wu, S.~Ge, S.~K. Zhou, and L.~Xiao, ``Knowledge matters: Chest radiology report generation with general and specific knowledge,'' \emph{Medical image analysis}, vol.~80, p. 102510, 2022.

\bibitem{21huang2023kiut}
Z.~Huang, X.~Zhang \emph{et~al.}, ``Kiut: Knowledge-injected u-transformer for radiology report generation,'' in \emph{CVPR}, 2023, pp. 19\,809--19\,818.

\bibitem{li2023DCL}
B.~Li, Mingjie~Lin \emph{et~al.}, ``Dynamic graph enhanced contrastive learning for chest x-ray report generation,'' in \emph{CVPR}, 2023, pp. 3334--3343.

\bibitem{17tanida2023RGRG}
T.~Tanida, P.~M{\"u}ller \emph{et~al.}, ``Interactive and explainable region-guided radiology report generation,'' in \emph{CVPR}, 2023, pp. 7433--7442.

\bibitem{15hou2023organ}
W.~Hou, K.~Xu, Y.~Cheng \emph{et~al.}, ``Organ: Observation-guided radiology report generation via tree reasoning,'' in \emph{ACL}, 2023, pp. 8108--8122.

\bibitem{16jin2024promptmrg}
H.~Jin, H.~Che \emph{et~al.}, ``Promptmrg: Diagnosis-driven prompts for medical report generation,'' in \emph{AAAI}, vol.~38, no.~3, 2024, pp. 2607--2615.

\bibitem{wu2chestImaGenome}
J.~T. Wu, N.~N. Agu \emph{et~al.}, ``Chest imagenome dataset for clinical reasoning,'' in \emph{Thirty-fifth Conference on Neural Information Processing Systems Datasets and Benchmarks Track (Round 2)}, 2021.

\bibitem{27johnson2019mimic}
A.~E. Johnson \emph{et~al.}, ``Mimic-cxr-jpg, a large publicly available database of labeled chest radiographs,'' \emph{arXiv preprint arXiv:1901.07042}, 2019.

\bibitem{10yang2021writing}
X.~Yang, M.~Ye \emph{et~al.}, ``Writing by memorizing: Hierarchical retrieval-based medical report generation,'' in \emph{ACL}, 2021, pp. 5000--5009.

\bibitem{xue2018rnn1}
Y.~Xue \emph{et~al.}, ``Multimodal recurrent model with attention for automated radiology report generation,'' in \emph{MICCAI}, 2018, pp. 457--466.

\bibitem{09xie2019attention}
X.~Xie \emph{et~al.}, ``Attention-based abnormal-aware fusion network for radiology report generation,'' in \emph{Database Systems for Advanced Applications: DASFAA 2019 International Workshops}, 2019, pp. 448--452.

\bibitem{12miura2021improving}
Y.~Miura, Y.~Zhang, E.~Tsai, C.~Langlotz, and D.~Jurafsky, ``Improving factual completeness and consistency of image-to-text radiology report generation,'' in \emph{ACL}, 2021, pp. 5288--5304.

\bibitem{13liu2021exploring}
F.~Liu \emph{et~al.}, ``Exploring and distilling posterior and prior knowledge for radiology report generation,'' in \emph{CVPR}, 2021, pp. 13\,753--13\,762.

\bibitem{ramesh2022hallucination}
V.~Ramesh, N.~A. Chi, and P.~Rajpurkar, ``Improving radiology report generation systems by removing hallucinated references to non-existent priors,'' in \emph{Machine Learning for Health}.\hskip 1em plus 0.5em minus 0.4em\relax PMLR, 2022, pp. 456--473.

\bibitem{he2016deep}
K.~He, X.~Zhang, S.~Ren, and J.~Sun, ``Deep residual learning for image recognition,'' in \emph{CVPR}, 2016, pp. 770--778.

\bibitem{liu2021swin}
Z.~Liu, Y.~Lin, Y.~Cao \emph{et~al.}, ``Swin transformer: Hierarchical vision transformer using shifted windows,'' in \emph{ICCV}, 2021, pp. 10\,012--10\,022.

\bibitem{smit2020CheXbert}
A.~Smit, S.~Jain, P.~Rajpurkar, A.~Pareek, A.~Y. Ng, and M.~Lungren, ``Combining automatic labelers and expert annotations for accurate radiology report labeling using bert,'' in \emph{EMNLP}, 2020, pp. 1500--1519.

\bibitem{devlin2019bert}
J.~Devlin, M.-W. Chang \emph{et~al.}, ``Bert: Pre-training of deep bidirectional transformers for language understanding,'' in \emph{ACL: Human Language Technologies, Volume 1 (Long and Short Papers)}, 2019, pp. 4171--4186.

\bibitem{r2gengpt}
Z.~Wang, L.~Liu \emph{et~al.}, ``R2gengpt: Radiology report generation with frozen llms,'' \emph{Meta-Radiology}, vol.~1, no.~3, p. 100033, 2023.

\bibitem{huang2019AoANet}
L.~Huang, W.~Wang, J.~Chen, and X.~Wei, ``Attention on attention for image captioning,'' in \emph{ICCV}, 2019, pp. 4634--4643.

\bibitem{moon2022cnntrans}
J.~H. Moon, H.~Lee, W.~Shin, Y.-H. Kim, and E.~Choi, ``Multi-modal understanding and generation for medical images and text via vision-language pre-training,'' \emph{IEEE Journal of Biomedical and Health Informatics}, vol.~26, no.~12, pp. 6070--6080, 2022.

\bibitem{CVT2Dis}
A.~Nicolson, J.~Dowling, and B.~Koopman, ``Improving chest x-ray report generation by leveraging warm starting,'' \emph{Artificial intelligence in medicine}, vol. 144, p. 102633, 2023.

\bibitem{14yang2023radiology}
S.~Yang, X.~Wu, S.~Ge, Z.~Zheng, S.~K. Zhou, and L.~Xiao, ``Radiology report generation with a learned knowledge base and multi-modal alignment,'' \emph{Medical Image Analysis}, vol.~86, p. 102798, 2023.

\bibitem{yang2022lavt}
Z.~Yang, J.~Wang \emph{et~al.}, ``Lavt: Language-aware vision transformer for referring image segmentation,'' in \emph{CVPR}, 2022, pp. 18\,155--18\,165.

\bibitem{2017Transformer}
A.~Vaswani, N.~Shazeer, N.~Parmar, J.~Uszkoreit, L.~Jones \emph{et~al.}, ``Attention is all you need,'' in \emph{NeurIPS}, 2017, p. 6000–6010.

\bibitem{xie2024fusionmamba}
X.~Xie, Y.~Cui, T.~Tan, X.~Zheng, and Z.~Yu, ``Fusionmamba: Dynamic feature enhancement for multimodal image fusion with mamba,'' \emph{Visual Intelligence}, vol.~2, no.~1, p.~37, 2024.

\bibitem{huang2022LDConv}
P.~Huang, H.~Ni, Y.~Ni \emph{et~al.}, ``Learnable descriptive convolutional network for face anti-spoofing.'' in \emph{BMVC}, vol.~2, no.~6, 2022, p.~7.

\bibitem{iuxray}
D.~DemnerFushman \emph{et~al.}, ``Preparing a collection of radiology examinations for distribution and retrieval,'' \emph{Journal of the American Medical Informatics Association}, vol.~23, no.~2, pp. 304--310, 2016.

\bibitem{papineni2002bleu}
K.~Papineni, S.~Roukos \emph{et~al.}, ``Bleu: a method for automatic evaluation of machine translation,'' in \emph{ACL}, 2002, pp. 311--318.

\bibitem{banerjee2005meteor}
S.~Banerjee and A.~Lavie, ``Meteor: An automatic metric for mt evaluation with improved correlation with human judgments,'' in \emph{Proceedings of the acl workshop on intrinsic and extrinsic evaluation measures for machine translation and/or summarization}, 2005, pp. 65--72.

\bibitem{lin2004rouge}
C.-Y. Lin, ``Rouge: A package for automatic evaluation of summaries,'' in \emph{Text summarization branches out}, 2004, pp. 74--81.

\bibitem{paszke2019pytorch}
A.~Paszke, S.~Gross, F.~Massa \emph{et~al.}, ``Pytorch: An imperative style, high-performance deep learning library,'' \emph{NeurIPS}, vol.~32, 2019.

\bibitem{loshchilov2017adamw}
I.~Loshchilov and F.~Hutter, ``Decoupled weight decay regularization,'' in \emph{ICLR}, 2019.

\bibitem{blip}
J.~Li, D.~Li, C.~Xiong, and S.~Hoi, ``Blip: Bootstrapping language-image pre-training for unified vision-language understanding and generation,'' in \emph{ICML}.\hskip 1em plus 0.5em minus 0.4em\relax PMLR, 2022, pp. 12\,888--12\,900.

\bibitem{wu2023RadFM}
C.~Wu, X.~Zhang, Y.~Zhang, Y.~Wang, and W.~Xie, ``Towards generalist foundation model for radiology by leveraging web-scale 2d\&3d medical data,'' \emph{arXiv preprint arXiv:2308.02463}, 2023.

\bibitem{li2024llavamed}
C.~Li, C.~Wong \emph{et~al.}, ``Llava-med: Training a large language-and-vision assistant for biomedicine in one day,'' \emph{NeurIPS}, vol.~36, 2024.

\bibitem{moor2023medflamingo}
M.~Moor \emph{et~al.}, ``Med-flamingo: a multimodal medical few-shot learner,'' in \emph{Machine Learning for Health (ML4H)}.\hskip 1em plus 0.5em minus 0.4em\relax PMLR, 2023, pp. 353--367.

\bibitem{he2024meddr}
S.~He, Y.~Nie, Z.~Chen, Z.~Cai, H.~Wang, S.~Yang, and H.~Chen, ``Meddr: Diagnosis-guided bootstrapping for large-scale medical vision-language learning,'' \emph{arXiv preprint arXiv:2404.15127}, 2024.

\bibitem{vlci}
W.~Chen, Y.~Liu, C.~Wang \emph{et~al.}, ``Cross-modal causal intervention for medical report generation,'' \emph{arXiv preprint arXiv:2303.09117}, 2023.

\bibitem{radialog}
C.~Pellegrini, E.~{\"O}zsoy, B.~Busam, N.~Navab, and M.~Keicher, ``Radialog: A large vision-language model for radiology report generation and conversational assistance,'' \emph{arXiv preprint arXiv:2311.18681}, 2023.

\bibitem{llmcxr}
S.~Lee, W.~Kim, J.~Chang, and J.~Ye, ``Llm-cxr: Instruction-finetuned llm for cxr image understanding and generation,'' in \emph{ICLR}, 2024.

\bibitem{maira2}
S.~Bannur, K.~Bouzid, D.~C. Castro \emph{et~al.}, ``Maira-2: Grounded radiology report generation,'' \emph{arXiv preprint arXiv:2406.04449}, 2024.

\bibitem{chexagent}
Z.~Chen, M.~Varma, J.-B. Delbrouck, M.~Paschali \emph{et~al.}, ``Chexagent: Towards a foundation model for chest x-ray interpretation,'' in \emph{AAAI 2024 Spring Symposium on Clinical Foundation Models}.

\bibitem{medversa}
H.-Y. Zhou and Aothers, ``Medversa: A generalist foundation model for medical image interpretation,'' \emph{arXiv preprint arXiv:2405.07988}, 2024.

\bibitem{tan2021efficientnetv2}
M.~Tan and Q.~Le, ``Efficientnetv2: Smaller models and faster training,'' in \emph{ICML}.\hskip 1em plus 0.5em minus 0.4em\relax PMLR, 2021, pp. 10\,096--10\,106.

\bibitem{rexrank}
X.~Zhang, H.-Y. Zhou, X.~Yang \emph{et~al.}, ``Rexrank: A public leaderboard for ai-powered radiology report generation,'' in \emph{AAAI Bridge Program on AI for Medicine and Healthcare}.\hskip 1em plus 0.5em minus 0.4em\relax PMLR, 2025, pp. 90--99.

\bibitem{bertscore}
T.~Zhang, V.~Kishore, F.~Wu, K.~Q. Weinberger, and Y.~Artzi, ``Bertscore: Evaluating text generation with bert,'' in \emph{ICLR}, 2020.

\bibitem{green}
S.~Ostmeier, J.~Xu, Z.~Chen \emph{et~al.}, ``Green: Generative radiology report evaluation and error notation,'' in \emph{EMNLP}, 2024, pp. 374--390.

\bibitem{yu2023RadCliQ}
F.~Yu, M.~Endo, R.~Krishnan \emph{et~al.}, ``Evaluating progress in automatic chest x-ray radiology report generation,'' \emph{Patterns}, vol.~4, no.~9, 2023.

\bibitem{gradcam}
R.~R. Selvaraju \emph{et~al.}, ``Grad-cam: Visual explanations from deep networks via gradient-based localization,'' in \emph{ICCV}, 2017, pp. 618--626.

\end{thebibliography}

\end{document}